\documentclass[journal]{IEEEtran}
\usepackage{comment}
\usepackage{lineno,hyperref}
\modulolinenumbers[5]
\usepackage{xfrac}
\usepackage{changepage}
\usepackage{amsfonts}
\usepackage{amsmath,amssymb}
\usepackage{graphicx}
\usepackage{subfigure}
\usepackage{multirow}
\usepackage{multicol}
\usepackage{array}
\newcolumntype{P}[1]{>{\centering\arraybackslash}p{#1}}
\usepackage{makecell}
\usepackage{pgfplots}
\usepackage[flushleft]{threeparttable}
\usepackage{alphalph}
\usepackage{bm}
\usepackage{cite}
\usepackage{pifont}
\usepackage{url}
\usepackage[figuresright]{rotating}
\definecolor{Lblue}{HTML}{18B8B4}
\definecolor{Lred}{HTML}{E74C3C}
\newcommand{\cmark}{\textcolor{green!80!black}{\ding{51}}}
\newcommand{\xmark}{\textcolor{red}{\ding{55}}}

\pgfplotsset{compat=1.16}
\begin{document}

\title{MultiScene: A Large-scale Dataset and Benchmark for Multi-scene Recognition in Single Aerial Images}

\author{Yuansheng~Hua,~
Lichao~Mou,~
Pu~Jin,
and~Xiao~Xiang~Zhu,~\IEEEmembership{Fellow, IEEE}

\thanks{The work is jointly supported by the European Research Council (ERC) under the European Union's Horizon 2020 research and innovation programme (grant agreement No. [ERC-2016-StG-714087], Acronym: \textit{So2Sat}), by the Helmholtz Association
through the Framework of Helmholtz AI (grant  number:  ZT-I-PF-5-01) - Local Unit ``Munich Unit @Aeronautics, Space and Transport (MASTr)'' and Helmholtz Excellent Professorship ``Data Science in Earth Observation - Big Data Fusion for Urban Research''(grant number: W2-W3-100) and by the German Federal Ministry of Education and Research (BMBF) in the framework of the international future AI lab "AI4EO -- Artificial Intelligence for Earth Observation: Reasoning, Uncertainties, Ethics and Beyond" (grant number: 01DD20001).
\textit{(Corresponding authors: Lichao Mou and Xiao Xiang Zhu.)}

Y. Hua, L. Mou, and X. X. Zhu are with the Remote Sensing Technology Institute, German Aerospace Center, 82234 Weßling, Germany, and also with the Data Science in Earth Observation, Technical University of Munich, 80333 Munich, Germany. (e-mails: yuansheng.hua@dlr.de; lichao.mou@dlr.de; xiaoxiang.zhu@dlr.de)

P. Jin is with the Data Science in Earth Observation, Technical University of Munich, 80333 Munich, Germany. (e-mail: pu.jin@tum.de)}}

\newpage
\thispagestyle{empty}
\onecolumn
\noindent This work has been submitted to the IEEE for possible publication. Copyright may be transferred without notice, after which this version may no longer be accessible.
\newpage
\twocolumn

\markboth{Submitted to IEEE TRANSACTIONS ON GEOSCIENCE AND REMOTE SENSING}%
{Hua \MakeLowercase{\textit{et al.}}: }

\maketitle
\begin{abstract}
\textcolor{blue}{This is the preprint version. To read the final version, please go to IEEE Transactions on Geoscience and Remote Sensing.} Aerial scene recognition is a fundamental research problem in interpreting high-resolution aerial imagery. Over the past few years, most studies focus on classifying an image into one scene category, while in real-world scenarios, it is more often that a single image contains multiple scenes. Therefore, in this paper, we investigate a more practical yet underexplored task---multi-scene recognition in single images. To this end, we create a large-scale dataset, called MultiScene, composed of 100,000 unconstrained high-resolution aerial images. Considering that manually labeling such images is extremely arduous, we resort to low-cost annotations from crowdsourcing platforms, e.g., OpenStreetMap (OSM). However, OSM data might suffer from incompleteness and incorrectness, which introduce noise into image labels. To address this issue, we visually inspect 14,000 images and correct their scene labels, yielding a subset of cleanly-annotated images, named MultiScene-Clean. With it, we can develop and evaluate deep networks for multi-scene recognition using clean data. Moreover, we provide crowdsourced annotations of all images for the purpose of studying network learning with noisy labels. We conduct experiments with extensive baseline models on both MultiScene-Clean and MultiScene to offer benchmarks for multi-scene recognition in single images and learning from noisy labels for this task, respectively. To facilitate progress, we make our dataset and trained models available on \url{https://gitlab.lrz.de/ai4eo/reasoning/multiscene}.

\end{abstract}


\IEEEpeerreviewmaketitle

\section{Introduction}
\label{sec:intro}

\begin{figure}
\centering
\subfigure[Single-scene recognition]{\includegraphics[width=.1878\textwidth]{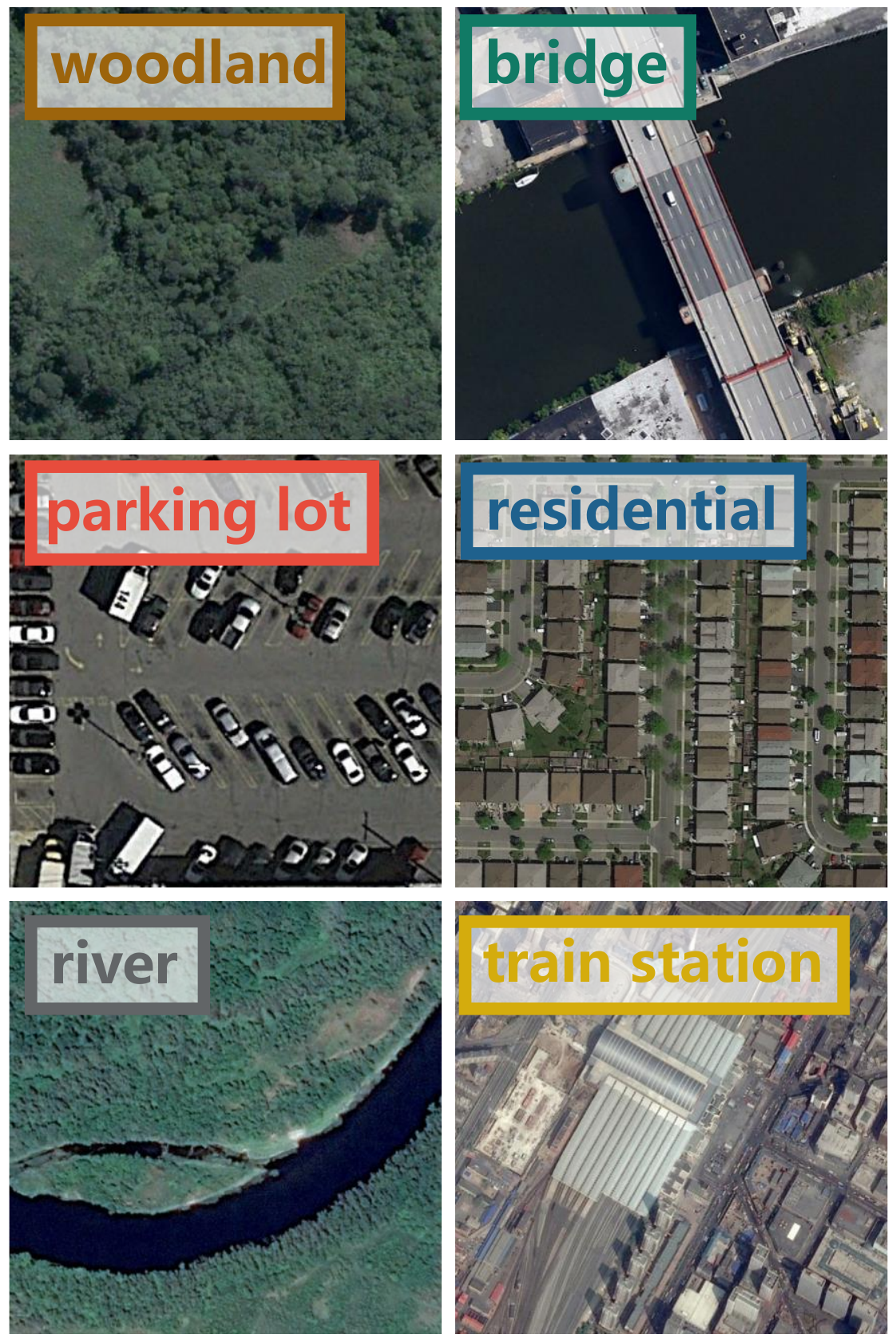}\label{fig:single}}
\subfigure[Multi-scene recognition]{\includegraphics[width=.283\textwidth]{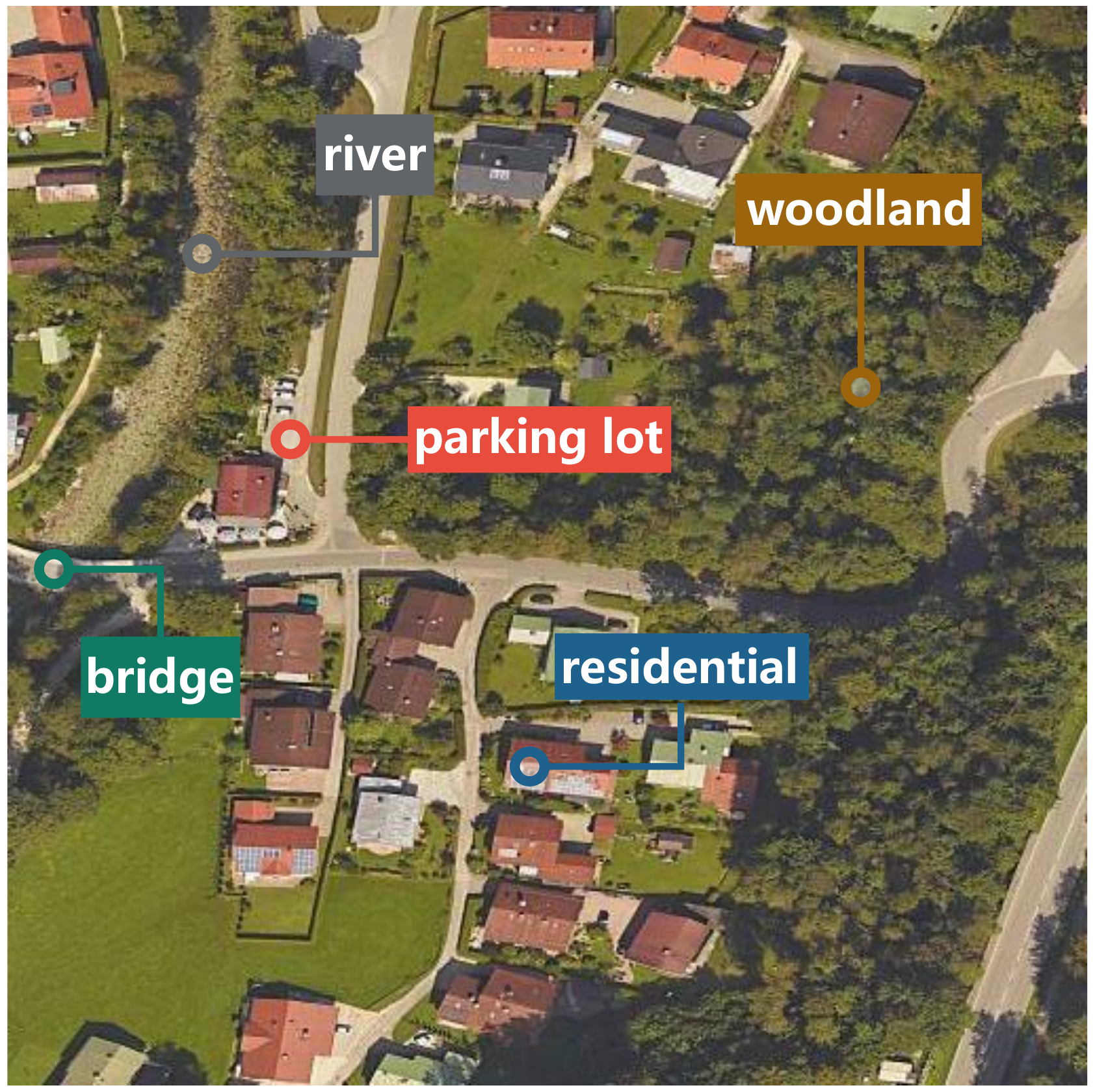}\label{fig:multi}}
\caption{Examples of images utilized in (a) single-scene and (b) multi-scene recognition tasks. In (a), each aerial image is assigned one scene label, while in (b), labels of all present scenes are inferred. In comparison with (b), (a) might suffer from partial scene understanding, as only one label is predicted even if there indeed exist multiple scenes in an image. For a clear visualization, locations of scenes are marked in (b).}
\label{fig:single_multi}
\end{figure}

With the recent development of Earth observation techniques, massive aerial imagery is now accessible for a variety of applications, such as environmental monitoring~\cite{weng2018land,cheng2017remote,wen2017semantic,manfreda2018use,mou18im,qiu2019local}, urban planning~\cite{marmanis17classification,beyongrgb,mou2018rifcn,li2017hsf,qiu2020fcn,li2020building}, land cover and land use mapping~\cite{cheng2013automatic,zhu2016bag,cheng2018deep,marcos18land}, and disaster assessment~\cite{vetrivel2018disaster, lee2017deep}. As one of the crucial steps towards these applications, aerial scene recognition has been extensively studied in the remote sensing community. During the last few years, the emergence of deep convolutional neural networks (CNNs) pushed ahead research in this field, and enormous achievements~\cite{hu2020cross,sun2020remote,murray2019zoom,cheng2020remote,byju2020remote,xu2020assessing,ru2020multi,li2020mapping} have been obtained. Albeit successful, most existing scene classification researches only focus on a specific scenario, where an aerial image is assumed to include a single scene~\cite{aid,aid++,tuia2016multi,niazmardi2017multiple,bi2020multiple,lin2020dual,wang2019multi,zhu2019high}. Basically, these studies regard aerial scene recognition as a single-label classification problem and learn models on well-cropped single-scene aerial images~(see Fig.~\ref{fig:single}). However, in practical applications, an aerial image often contains multiple scenes, as it is collected overhead and usually has a large coverage~(cf. Fig.~\ref{fig:multi}). We also note that even in public single-scene aerial image datasets, the coexistence of multiple scenes in a single image is inevitable, especially in images covering large areas. For example, as shown in the bottom two images in Fig.~\ref{fig:single}, although they are assigned single scene labels according to their central/dominant scenes (i.e., river and train station), there actually exists more than one scene in each of them.

Hence, in this paper, we aim to tackle a more realistic yet challenging problem, namely multi-scene recognition in single aerial images. This task refers to assigning an aerial image multiple scene labels, and there are no constraints on image preparations, such as centering dominant scenes and eliminating clutter scenes. Compared to the conventional scene recognition task, multi-scene recognition is more arduous because 1) images are large-scale and unconstrained, and 2) all present scenes in an aerial image need to be exhaustively recognized. Fig.~\ref{fig:single_multi}(b) shows an example of multi-scene aerial image and corresponding multiple scene-level labels. We can see that not only dominant scenes (e.g., residential and woodland) but also trivial scenes (e.g., bridge and parking lot) are annotated, which draws a more comprehensive picture for the unconstrained image.

\begin{figure}
\centering
\subfigure[An example of incomplete OSM data]{\includegraphics[width=.48\textwidth]{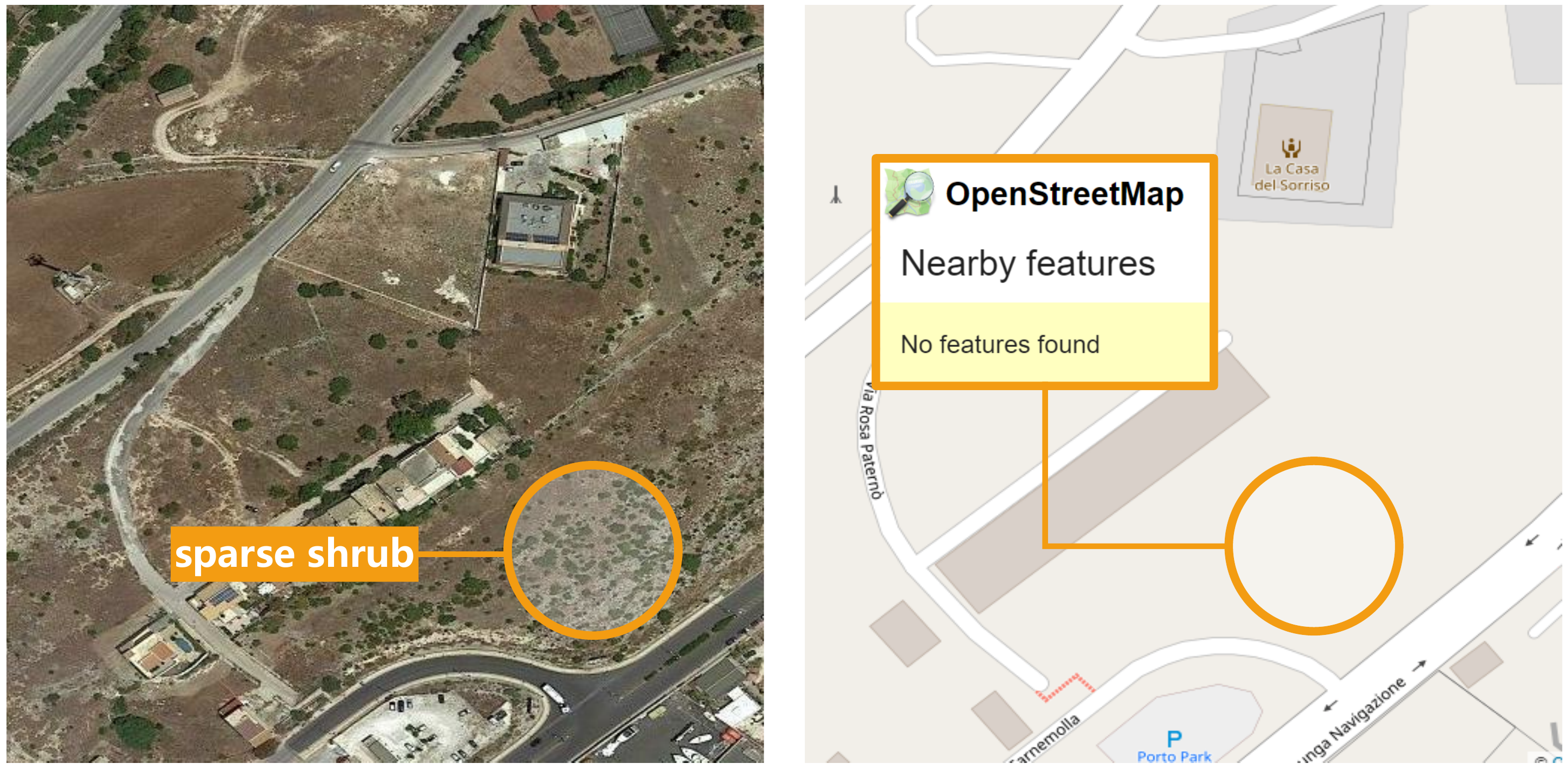}}
\subfigure[An example of incorrect OSM data]{\includegraphics[width=.48\textwidth]{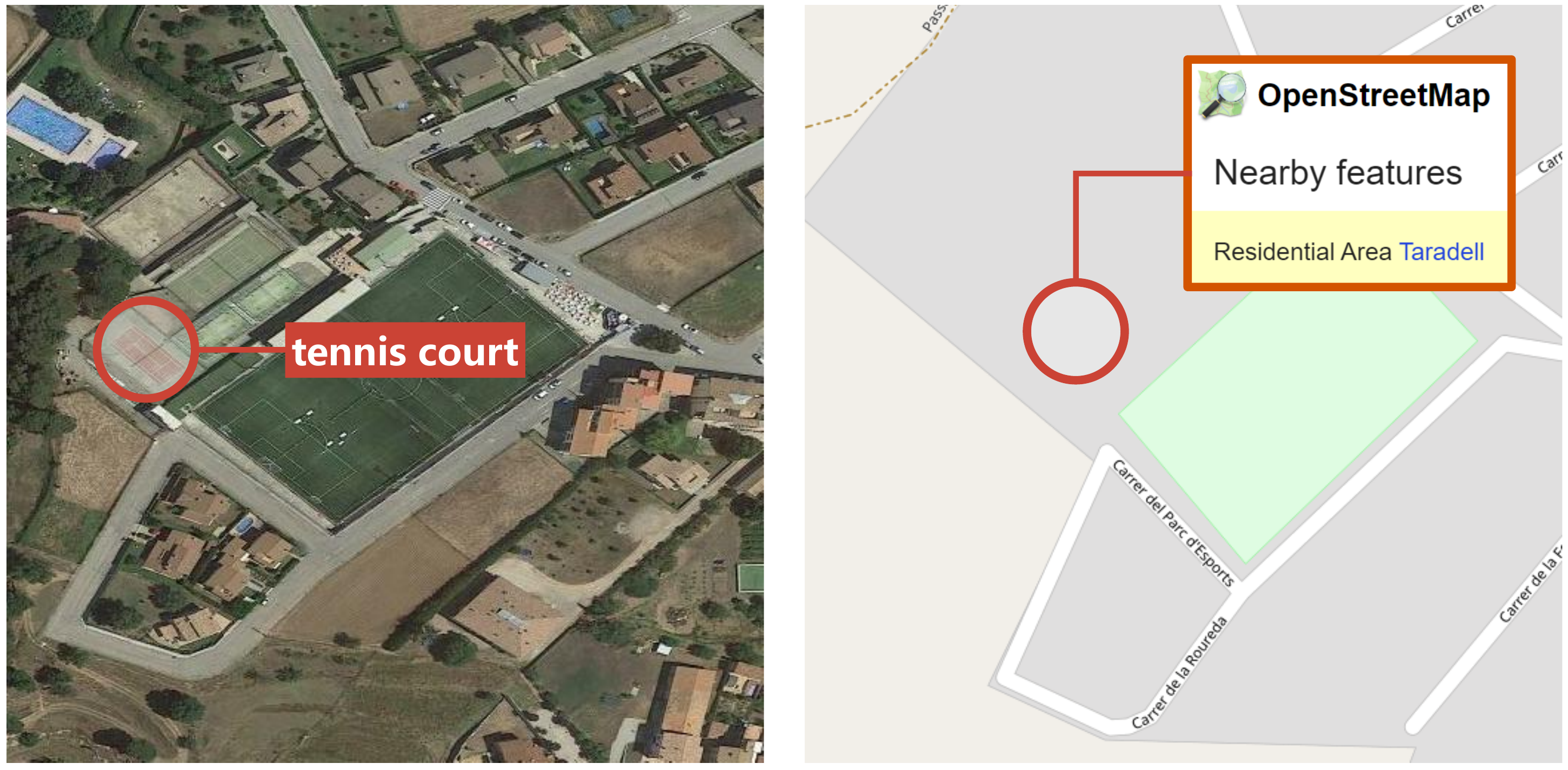}}
\caption{Examples of (a) incomplete and (b) incorrect OSM annotations. In (a), sparse shrubs are not annotated in OSM data, while in (b), the tennis court is mislabeled as residential.}
\label{fig:wrong_osm}
\end{figure}

However, very few efforts have been deployed to this problem in the remote sensing community. In order to advance the progress of multi-scene recognition in single images, we propose a large-scale Multi-Scene recognition (MultiScene) dataset, where 100,000 aerial images are collected around the world. In the phase of data preparation, we note that although massive high-resolution aerial images can be effortlessly obtained from remote sensing data platforms, such as Google Earth~\footnote{\url{https://earth.google.com/web/}}, it is extremely time- and labor-consuming to yield their corresponding multiple scene labels. To alleviate such annotation burden, in this paper, we resort to crowdsourced data, e.g., OpenStreetMap~\footnote{\url{https://www.openstreetmap.org/}} (OSM) annotations, which has been proven to be successful in generating image-level labels~\cite{aid,aid++,aid1m} and pixel-wise footprints~\cite{li2020building,zorzi2019regularization} for training deep networks. However, we observe that OSM data might suffer from two common defects, incompleteness and incorrectness, which could introduce severe noise into image labels. Fig~\ref{fig:wrong_osm} shows two examples of incorrect OSM annotations, where (a) sparse shrubs are neglected, and (b) the tennis court is mislabeled as residential. With this in mind, here we do not directly use crowdsourced labels as ground truth data. Instead, we visually inspect 14,000 images and correct their labels, producing a subset of cleanly-labeled images, named MultiScene-Clean. It allows developing and evaluating deep networks for unconstrained multi-scene recognition using clean data. Moreover, we note that the noisy crowdsourced data are not completely useless, for example, they can be used to study network learning with noisy labels for this task. Therefore, we also provide crowdsourced annotations of all images.

The contributions of this paper are four-fold:

\begin{table*}[!t]
\footnotesize
\centering
\caption{Comparison with existing aerial scene datasets from various perspectives.}
\label{tab:comparison_datasets}
\begin{tabular}{p{3cm}|>{\centering\arraybackslash}p{1.5cm}>{\centering\arraybackslash}p{2.5cm}>{\centering\arraybackslash}p{1.5cm}>{\centering\arraybackslash}p{2.5cm}>{\centering\arraybackslash}p{2.5cm}>{\centering\arraybackslash}p{0.5cm}}
\Xhline{3\arrayrulewidth}
\textbf{Dataset} & \textbf{\# images} & \textbf{spatial resolutions} & \textbf{\# scenes} & \textbf{\# labels per image} & \textbf{crowdsourced label} & \textbf{Year} \\
\hline
UC-Merced~\cite{ucm} & 2,100 & 0.3 m/pixel & 21 & 1 & \xmark & 2010 \\
WHU20~\cite{whu20} & 5,000 & 0.3-7.4 m/pixel & 20 & 1 & \xmark & 2015 \\
RSSCN7~\cite{rsscn7} & 2,800 & 0.2-1.4 m/pixel & 7 & 1 & \xmark & 2015 \\
AID~\cite{aid} & 10,000 & 0.5-8 m/pixel & 30 & 1 & \xmark & 2017 \\
NWPU-RESISC45~\cite{nwpu} & 31,500 & 0.2-30 m/pixel & 45 & 1 & \xmark & 2017\\
\textbf{MultiScene (Ours)} & 100,000 & 0.3-0.6 m/pixel & 36 & 1-13 & \cmark & 2021\\
\Xhline{3\arrayrulewidth}
\end{tabular}
\end{table*}

\begin{itemize}
\item Unlike conventional aerial scene recognition where all images are well-cropped and each of them contains only one scene-level label, in this paper, we explore a more practical task---multi-scene recognition in single images.
\item We propose a large-scale dataset, namely MultiScene, consisting of 100,000 unconstrained multi-scene aerial images, and each is assigned OSM labels. We visually inspect 14,000 images and correct their labels, yielding a subset of cleanly-labeled images.
\item The proposed dataset provides not only ground truth data but also crowdsourced labels, which enables researches in learning from enormous noisy labels for our task. 
\item We extensively evaluate commonly-used classification networks on both MultiScene-Clean and MultiScene and provide benchmarks for recognizing multiple scenes in single images and learning from noisy labels for this task, respectively.
\end{itemize}

The remaining sections of this paper are organized as follows. Section~\ref{sec:review} reviews studies in aerial single-scene classification and multi-label object classification. Section~\ref{sec:dataset} briefly recalls existing scene datasets and delineates the proposed dataset. Experimental configurations and results are exhibited in Section~\ref{sec:experiments}, and Section~\ref{sec:conclusion} draws a conclusion.

\section{Related Work}
\label{sec:review}

This section briefly reviews related works in two fields: aerial single-scene classification and multi-label object recognition.

\subsection{Aerial Single-scene Classification}

Aerial single-scene classification refers to categorize an aerial image into a single scene class. Early researches propose to construct scene representations with variant low-level features, e.g., local structures~\cite{sift,gabor}, color attributes~\cite{color1,color2}, and texture information~\cite{texture,texture_similar}. Concerning that low-level features fail to comprehensively depict complex scenes, mid-level algorithms, such as Bag-of-Visual-Words (BoVW)~\cite{bovw1,bovw2} and topic models~\cite{topic1,topic2}, are devised to encode local features (so-called ``visual words'') into more holistic mid-level scene representations for the classification task. However, these methods show limited performance in recognizing scenes of high diversity due to their dependency on hand-crafted features.

Recently, the emergence of deep CNNs brings immense advancements to the community, and many achievements~\cite{aid,aid++,hu2020cross,sun2020remote,tuia2016multi,murray2019zoom,cheng2020remote,bi2020multiple,byju2020remote,niazmardi2017multiple,xu2020assessing,ru2020multi,li2020mapping,wang2019multi,lin2020dual,zhu2019high} have been obtained in the field of aerial single-scene classification. These deep networks have hierarchical architectures, where convolutional and max-pooling layers are periodically interleaved for learning high-level features of intricate scenes. With layers going deeper, the learned features are more abstract and supposed to contain richer semantic information, which is crucial for judicious decisions. A popular trend of deep learning algorithms in single-scene classification is to take a CNN as the backbone and introduce well-designed modules for further enhancing the feature efficiency. For instance, Bi et al.~\cite{bi2020multiple} propose to learn multiple instances from feature maps extracted by a densely-connected CNN and integrate them into bag-level features for single-scene classification. Li et al.~\cite{li2020kfb} propose a key region capturing method to learn class-specific features and retain global information for inferring scene labels. To leverage features of variant levels, feature aggregation plays a key role in single-scene classification. Lu et al.~\cite{lu2019faccnn} fuses features learned by the last three blocks and the second fully-connected layer of VGG-16, and Cao et al.~\cite{cao2020saff} designs a non-parametric self-attention layer to enhance spatial and channel responses of fused features for the final prediction. In~\cite{sun2020remote}, the authors develop a gated bidirectional network for aggregating features extracted by different convolutional layers with a gated function in both top-down and bottom-up directions. Besides, exploiting supplementary data, such as geo-tagged audios and multi-temporal images, has been a new research direction. Hu et al.~\cite{hu2020cross} propose to predict scene categories by transferring sound event knowledge learned from sound-image pairs. In~\cite{ru2020multi}, the authors propose a two-branch network to learn deep features of bi-temporal images and fuse them through a CorrFusion module for aerial scene classification. Our literature review demonstrates that most of the existing researches assume that an aerial image includes only one scene and focus on well-cropped single-scene aerial images. Hence, these studies tend to regard entities present in an image as compositions of a scene, while in multi-scene recognition, this would trigger networks to learn erroneous feature representations. However, very few efforts have been deployed to explore multi-scene recognition in the remote sensing community.

\begin{figure*}
\centering
\includegraphics[width=.98\textwidth]{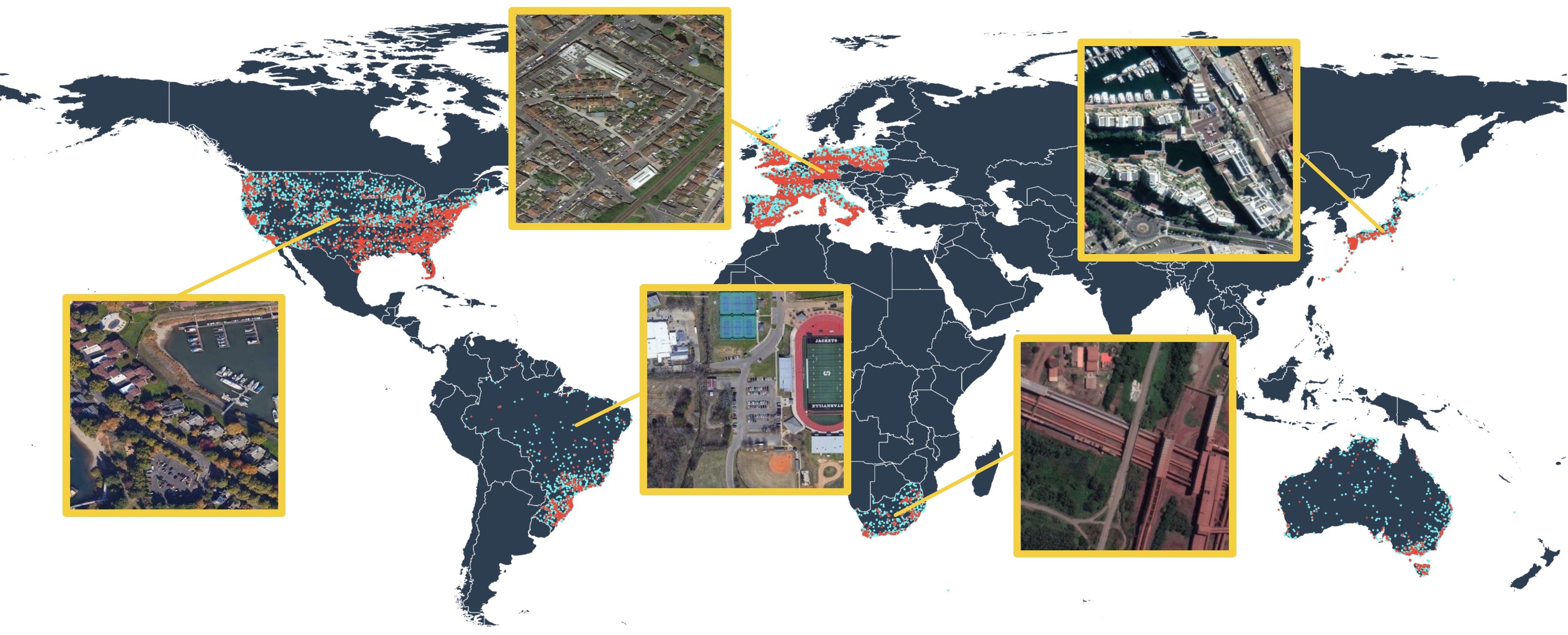}
\caption{Coordinate distributions and examples of multi-scene aerial images in our dataset. \textcolor{Lred}{Red} dots denote images with both crowdsourced and clean labels, and \textcolor{Lblue}{cyan} dots represent images with only crowdsourced scene labels.}
\label{fig:data_distribution}
\end{figure*}

\subsection{Multi-label Object Classification}
Multi-label object classification refers to assigning an aerial image multiple object-level labels, such as car, tree, and building. Similar to our work, these studies aim to provide a holistic understanding of aerial images, but from the perspective of object. Early attempts~\cite{zeggada2017deep,zeggada2018multilabel} follow the idea of simply combining a deep CNN with a post-processing approach for identifying multiple objects in an aerial image. In~\cite{zeggada2017deep}, the authors feed outputs of a CNN into a customized thresholding operation for inferring multiple object labels, while in~\cite{zeggada2018multilabel}, a conditional random field (CRF) is utilized as the post-processing model. In recent literature, more efforts are deployed to endow deep neural networks with the capacity of reasoning about relations among various objects for more accurate predictions. In~\cite{hua2019recurrently}, the authors propose an end-to-end network comprising a CNN and a long short-term memory (LSTM) network that is responsible for modeling label dependencies through its recurrent units for multi-label object classification. \cite{sumbul2020attention} exploits a bidirectional LSTM network to learn spatial relations among all patches in an image for the final prediction. In \cite{hua2020relation}, the authors propose a relational reasoning network module to model label dependencies and gains better classification results. Instead of encoding label relations, \cite{koda2018spatial} divides an aerial image into several patches with the same size and models spatial relationships among them for multi-label object interpretation. Compared to these researches, our task is more challenging, because compared to object, the concept of scene is more abstract and intricate.

\section{MultiScene Dataset for Multi-Scene Recognition in Single Aerial Images}
\label{sec:dataset}
This section first reviews existing single-scene aerial image datasets and then delineates the proposed dataset.

\subsection{Existing Single-scene Aerial Image Dataset}

During the last decades, various aerial image datasets are published for single-scene classification, and here we briefly review several commonly used ones.

\begin{itemize}
\item \textit{UC-Merced~\cite{ucm}:} The UC-Merced dataset is composed of 2,100 images collected from the United States Geological Survey (USGS) National Map, and each of them is categorized into one of 21 scene classes: overpass, golf course, river, harbor, beach, building, airplane, freeway, intersection, medium residential, runway, agricultural, storage tank, parking lot, forest, sparse residential, chaparral, tennis courts, dense residential, baseball diamond, and mobile home park. The number of images per scene is evenly defined as 100, and only cities in the United States are covered in data acquisition. The size of each image is $256 \times 256$ pixels, and the spatial resolution is one foot. In~\cite{ucm-mul}, the authors focus on the task of recognizing multiple objects in an image and relabel the UC-Merced dataset, yielding a multi-label dataset. In this dataset, 2,100 images are relabeled, and each is assigned one or several labels from 17 newly defined object classes: airplane, sand, pavement, building, car, chaparral, court, tree, dock, tank, water, grass, mobile home, ship, bare soil, sea, and field.

\item \textit{WHU20~\cite{whu20}:} The WHU20 dataset is an extended version of the WHU-RS dataset that was originally proposed in~\cite{whu10}. This dataset expands numbers of aerial images and scene classes from 950 to 5,000 and from 12 to 20, respectively. For each scene category, more than 200 images with a size of $600 \times 600$ pixels are collected, and their spatial resolutions range from 0.26 m/pixel to 7.44 m/pixel.

\item \textit{RSSCN7~\cite{rsscn7}:} The RSSCN7 dataset is a collection of 2,800 high-resolution images each belonging to one of 7 scene categories: grassland, forest, farmland, parking lot, river/lake, industrial region, and residential region. 400 images with different spatial resolutions are cropped from Google Earth imagery for each scene, and the image size is $400 \times 400$ pixels.

\item \textit{AID~\cite{aid}:} The AID dataset is a large-scale benchmark consisting of 10,000 aerial images and 30 scene types: airport, pond, forest, baseball field, resort, bare land, center, beach, bridge, commercial, desert, storage tanks, farmland, industrial, mountain, park, parking, playground, viaduct, church, railway station, river, school, meadow, sparse residential, dense residential, medium residential, square, stadium, and port. Google Earth is exploited to acquire image samples, and the spatial resolution of each sample varies from 0.5 m/pixel to 8 m/pixel. The size of images is $600 \times 600$ pixels, and the number of images for each class ranges from 220 to 420. 

\item \textit{NWPU-RESISC45~\cite{nwpu}:} The NWPU-RESISC45 dataset contains 31,500 high-resolution images and each is assigned with one of 45 scene labels. For each scene, 700 images with a size of $256 \times 256$ pixels are acquired from Google Earth imagery, and their spatial resolutions vary from 0.2 m/pixel to 30 m/pixel.

\end{itemize}

In addition, we note that BigEarthNet~\cite{bigearthnet} is a large-scale dataset for multi-label learning, where 590,326 Sentinel-2 images are captured over the European Union, and their spatial resolutions range from 10 m/pixel to 60 m/pixel. Since BigEarthNet focuses on land covers instead of scenes, we do not specify it here. Table~\ref{tab:comparison_datasets} presents an overview of public high-resolution aerial image datasets from the perspectives of dataset scales, image resolutions, scene categories, and annotations.

\subsection{MultiScene for Multi-scene Recognition}

\begin{figure*}
\centering
\includegraphics[width=.95\textwidth]{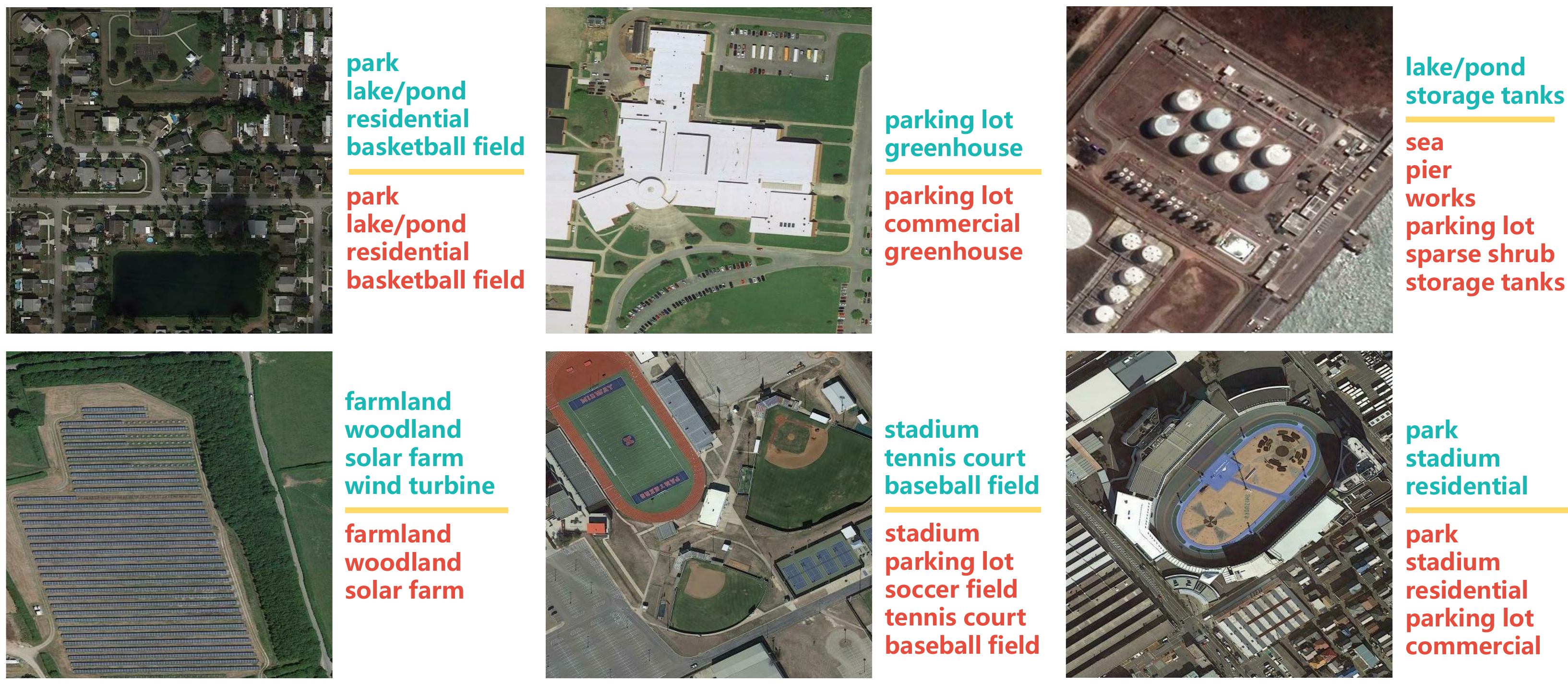}
\caption{Example multi-scene aerial images with their \textcolor{Lblue}{crowdsourced} and \textcolor{Lred}{clean} annotations in the MultiScene dataset.}
\label{fig:data_examples}
\end{figure*}

Although there are already variant datasets for aerial scene recognition, most of them can only be used for single-scene classification. In this paper, we aim to take a step towards a more general scenario, multi-scene recognition in single images, and produce the MultiScene dataset.

To be more specific, we collect 100,000 high-resolution aerial images from Google Earth imagery, which cover six continents, Europe, Asia, North America, South America, Africa, and Oceania, and eleven countries including Germany, France, Italy, England, Spain, Poland, Japan, the United States, Brazil, South Africa, and Australia (cf. Fig.~\ref{fig:data_distribution}). This can ensure high intra-class diversity, as different scene appearances resulted from different cultural regions are covered. The spatial resolution of each image ranges from 0.3 m/pixel to 0.6 m/pixel, and the spatial size of images is $512 \times 512$ pixels. In contrast to single-scene image datasets~\cite{aid,rsscn7,ucm,whu20}, we put no constraints on the location and area of the dominant/trivial scene in an image during the data collection process. Some example multi-scene images are exhibited in Fig.~\ref{fig:data_examples}. In total, 36 scene categories are defined: apron, baseball field, basketball field, beach, bridge, cemetery, commercial, farmland, woodland, golf course, greenhouse, helipad, lake/pond, oil field, orchard, parking lot, park, pier, port, quarry, railway, residential, river, roundabout, runway, soccer field, solar farm, sparse shrub, stadium, storage tanks, tennis court, train station, wastewater, plant, wind turbine, works, and sea.

To obtain crowdsourced annotations, we first localize each image in OSM with coordinates of its four corners. Afterwards, we parse properties of scenes present in the corresponding region and label images accordingly. In this way, crowdsourced annotations of all aerial images can be automatically yielded at a very low cost compared to conventional manual labeling. However, these almost free annotations might suffer from noise as aforementioned in Section~\ref{sec:intro}, and the performance of networks directly trained on them could be degraded. Therefore, we visually inspect 14,000 images
from all six continents and correct their labels, yielding a subset, MultiScene-Clean. Fig.~\ref{fig:data_distribution} shows the coordinate distribution of all images, and the number of samples associated with each scene is present in Fig.~\ref{fig:data_statistic}. Compared to other scene recognition datasets (cf. Table~\ref{tab:comparison_datasets}), our dataset is featured by its manifold labels per image and the available crowdsourced annotations. Fig.~\ref{fig:scene_number} further shows the number of images associated with different numbers of scenes.

\subsection{Challenges}
\begin{figure*}
\centering
\includegraphics[width=.98\textwidth]{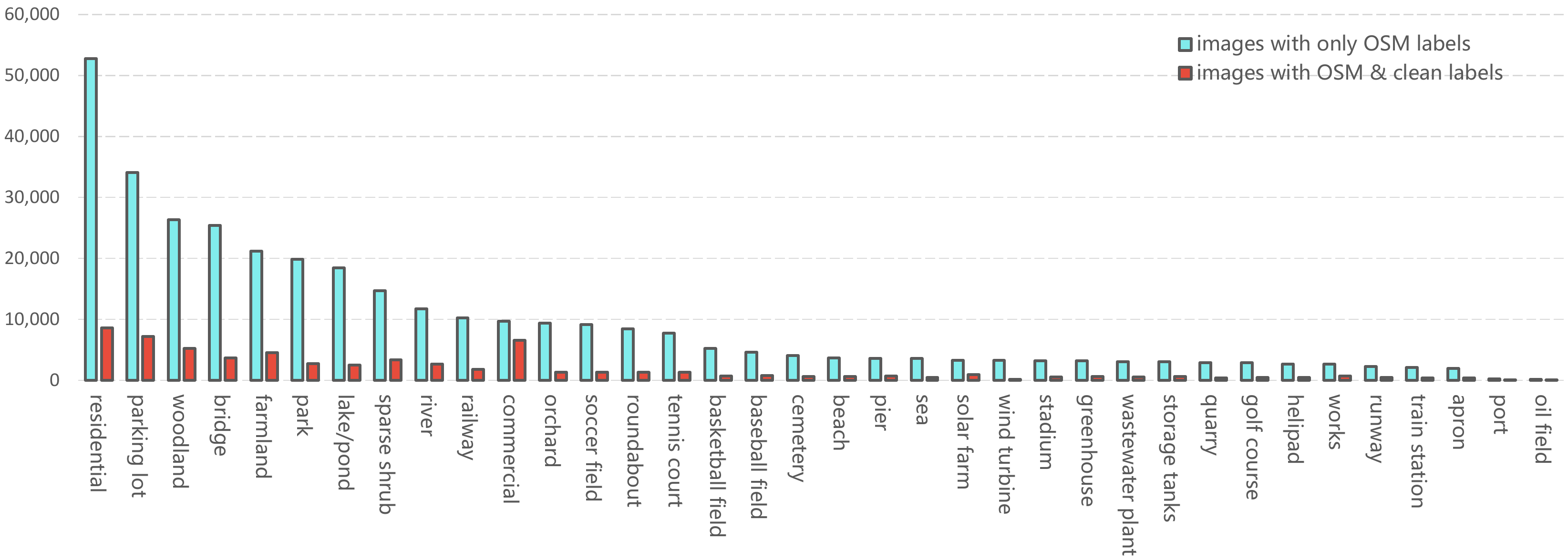}
\caption{Sample distributions of all scene categories in our dataset. Each \textcolor{Lblue}{cyan} bar indicates the number of images assigned only OSM labels with respect to each scene category, and \textcolor{Lred}{red} bars represent numbers of images with both OSM and clean labels.}
\label{fig:data_statistic}
\end{figure*}

\begin{figure}
\centering
\includegraphics[width=.45\textwidth]{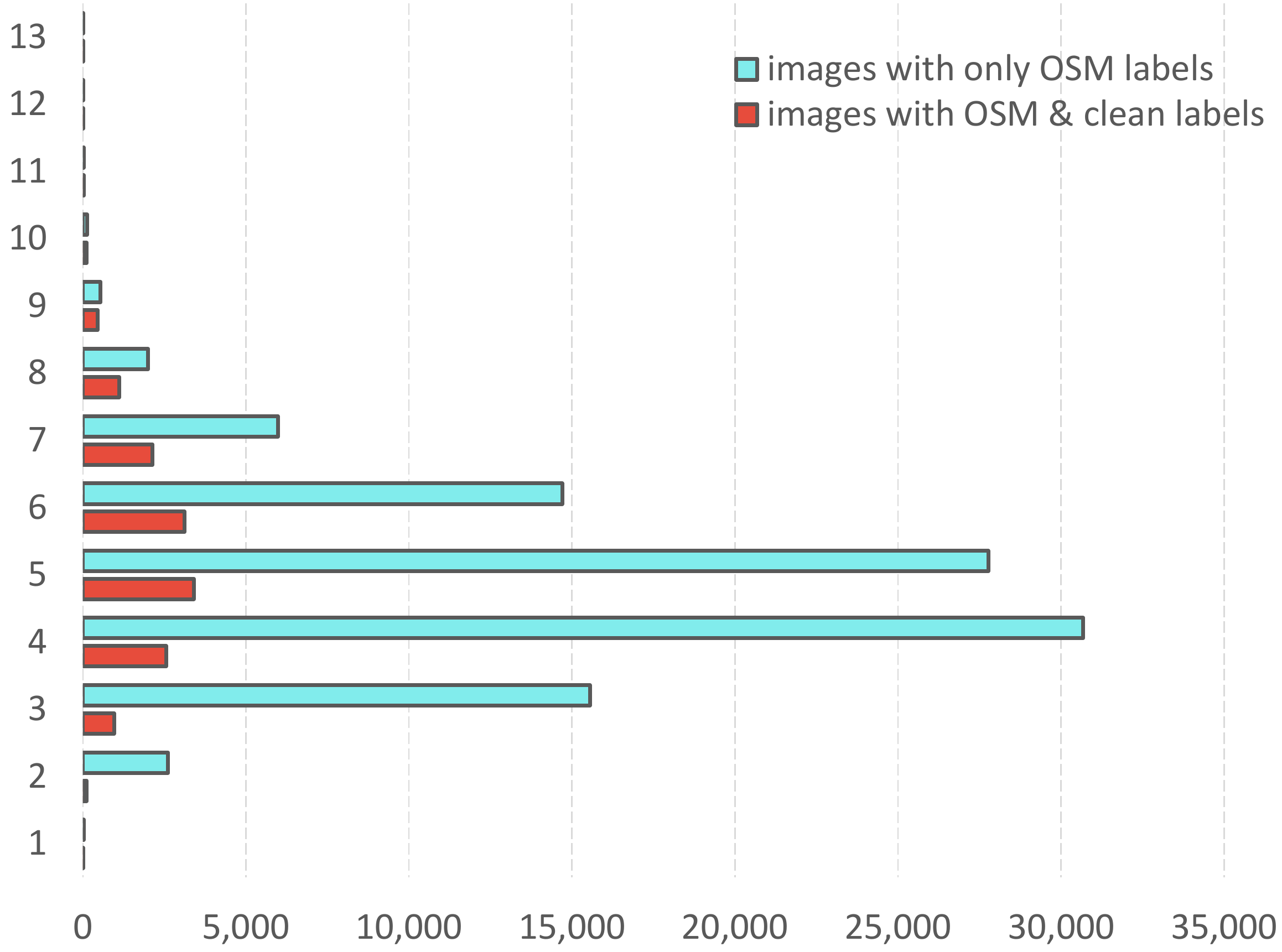}
\caption{The number of images associated with different numbers of scenes. Y-axis indicates the number of scenes, and X-axis represents the number of images. The legend is the same as that in Fig.~\ref{fig:data_statistic}.}
\label{fig:scene_number}
\end{figure}

Compared to existing aerial scene datasets, our dataset brings more challenges to the field of scene interpretation from the following three perspectives:

\begin{itemize}
\item Images are unconstrained and large-scale, and thus scenes are likely to be incomplete and trivial, which makes recognition more difficult.

\item The long-tail sample distribution (see Fig.~\ref{fig:data_statistic}) poses a challenge of learning unbiased models on an imbalanced dataset.

\item We gather images from different cultural regions, which results in a high intra-class variation.

\end{itemize}

\section{Experiments}
\label{sec:experiments}
\subsection{Experimental Setup}

\textbf{Data Configuration.} We evaluate the performance of existing models on both MultiScene-Clean and MultiScene datasets. As to the former, we use 7,000 cleanly labeled images to train and validate networks, and the remaining images are utilized to test networks. For the latter, we leverage the same test set but train deep neural networks on the other 93,000 images with only crowdsourced annotations.

\textbf{Evaluation.} For a comprehensive evaluation, we measure the performance of baseline models with class-based, example-based, and overall metrics. Let $L$ and $N$ be numbers of classes and examples\footnote{An example indicates an image which has multiple labels.}, these metrics are calculated as follows.
\begin{itemize}
\item \textit{Class-based Metrics:} Mean class-based precision (mCP), recall (mCR), F$_1$ (mCF$_1$) score, and per-class average precision (AP) are calculated for measuring the performance of networks from the perspective of class. Specifically,
mCP, mCR, and mCF$_1$ score are computed as:
\begin{equation}
\begin{gathered}
\text{mCP} = \frac{1}{L}\sum_{c=1}^{L}\frac{\text{TP}_c}{\text{TP}_c+\text{FP}_c},\; \text{mCR} = \frac{1}{L}\sum_{c=1}^{L}\frac{\text{TP}_c}{\text{TP}_c+\text{FN}_c}, \\\text{mCF}_1 = \frac{1}{L}\sum_{c=1}^{L}\frac{\text{TP}_c}{\text{TP}_c+\frac{1}{2}(\text{FP}_c+\text{FN}_c)},
\end{gathered}
\end{equation}
where TP$_c$, FN$_c$, and FP$_c$ represent numbers of true positives, false negatives, and false positives with respect to the $c$-th class, respectively. As to the per-class AP, we first rank all examples according to the predicted probability of the $c$-th class in each of them. Then we calculate the corresponding AP with the following formula:
\begin{equation}
\text{AP} = \frac{1}{N_c}\sum_{k=1}^{N}\frac{\text{TP}_c@k}{\text{TP}_c@k+\text{FP}_c@k}\times \text{rel}@k,
\end{equation}
where $N_c$ denotes the number of examples including the $c$-th class, and TP$_c@k$ and FP$_c@k$ represent numbers of true and false positives in top-$k$ examples, respectively. Notably, TP$_c@k$ and FP$_c@k$ are equivalent to TP$_c$ and FP$_c$, when $k$ equals to $N$. rel@$k$ denotes the relevance between the $k$-th example and the $c$-th class, and it is set to 0/1 when the $c$-th class is included/excluded. Besides, the mean average precision (mAP) can be computed by averaging APs for all categories.

\item \textit{Example-based Metrics:} Mean example-based precision (mEP), recall (mER), and F$_1$ (mEF$_1$) score are computed to validate networks from the perspective of example with the following equations:
\begin{equation}
\begin{gathered}
\text{mEP} = \frac{1}{N}\sum_{k=1}^{N}\frac{\text{TP}_k}{\text{TP}_k+\text{FP}_k},\; \text{mER} = \frac{1}{N}\sum_{k=1}^{N}\frac{\text{TP}_k}{\text{TP}_k+\text{FN}_k}, \\\text{mEF}_1 = \frac{1}{N}\sum_{k=1}^{N}\frac{\text{TP}_k}{\text{TP}_k+\frac{1}{2}(\text{FP}_k+\text{FN}_k)},
\end{gathered}
\end{equation}
where TP$_k$, FP$_k$, and FN$_k$ denote numbers of true positives, false positives, and false negatives in the $k$-th example.

\item \textit{Overall Metrics:} Overall precision (OP), recall (OR), and F$_1$ (OF$_1$) score can be used to measure the performance of models from a more holistic perspective, and they are calculated as:
\begin{equation}
\begin{gathered}
\text{OP} = \frac{\text{TP}}{\text{TP}+\text{FP}},\; \text{OR} = \frac{\text{TP}}{\text{TP}+\text{FN}}, \\\text{OF}_1 = \frac{\text{TP}}{\text{TP}+\frac{1}{2}(\text{FP}+\text{FN})},
\end{gathered}
\end{equation}
where TP, FP, and FN are counted based on predictions of all scenes and examples. 

\end{itemize}

\begin{table*}[t]
\footnotesize
\centering
\renewcommand{\arraystretch}{1}
\caption{Numerical results of baseline models on the MultiScene-Clean dataset (\%). Models are trained and tested on cleanly-labeld images, and the best scores are shown in bold.}
\label{tab:msrlite_overall_results}
\begin{tabular}{p{2cm}|*{4}{p{0.6cm}}|*{3}{p{0.6cm}}|*{3}{p{0.6cm}}}
\Xhline{3\arrayrulewidth}
\textbf{Model} & \textbf{mAP} & \textbf{mCP} & \textbf{mCR} & \textbf{mCF$_1$} & \textbf{mEP} & \textbf{mER} & \textbf{mEF$_1$} & \textbf{OP} & \textbf{OR} & \textbf{OF$_1$} \\
\hline

SVM & 14.9 & 19.6 & 8.4 & 8.6 & 62.2 & 32.8 & 41.1 & 66.9 & 32.2 & 43.5 \\
RF & 15.6 & 25.4 & 8.7 & 9.5 & 64.6 & 32.5 & 41.4 & 70.9 & 32.1 & 44.2 \\
XGBOOST & 16.9 & 34.1 & 11.2 & 12.8 & 67.0 & 37.4 & 45.8 & 69.6 & 36.5 & 47.9 \\
VGG-16 & 56.5 & 63.3 & 47.9 & 53.6 & 74.9 & 64.3 & 67.0 & 73.6 & 63.1 & 67.9 \\
VGG-19 & 56.4 & 62.9 & 47.7 & 53.3 & 74.8 & 64.1 & 66.8 & 73.5 & 62.7 & 67.7 \\
Inception-V3 & 53.5 & 65.0 & 40.8 & 48.5 & 74.2 & 59.9 & 63.9 & 73.0 & 58.6 & 65.0 \\
ResNet-50 & 62.0 & 74.8 & 45.9 & 55.1 & 79.7 & 62.7 & 67.9 & 79.0 & 61.4 & 69.1 \\
ResNet-101 & 63.0 & 75.9 & 46.6 & 55.8 & 79.9 & 64.3 & 69.1 & 79.2 & 63.1 & 70.3\\
ResNet-152 & 63.8 & 74.9 & 49.1 & 57.7 & \textbf{80.8} & 64.0 & 69.2 & \textbf{80.1} & 62.8 & 70.4 \\
SqueezeNet & 46.3 & 58.1 & 36.8 & 43.5 & 71.3 & 58.0 & 61.3 & 70.0 & 56.9 & 62.7 \\
MobileNet-V2 & 58.8 & 70.9 & 44.8 & 53.1 & 77.6 & 62.7 & 67.0 & 76.6 & 61.6 & 68.3\\
ShuffleNet-V2 & 50.7 & 61.8 & 38.1 & 45.7 & 73.8 & 58.2 & 62.5 & 73.0 & 57.0 & 64.0\\
DenseNet-121 & 62.2 & 74.6 & 45.1 & 54.4 & 79.5 & 61.8 & 67.3 & 79.1 & 60.6 & 68.6\\
DenseNet-169 & 63.2 & 76.7 & 45.8 & 55.3 & 80.4 & 63.4 & 68.6 & 79.6 & 62.3 & 69.9 \\
ResNeXt-50 & 63.4 & \textbf{77.3} & 45.0 & 54.2 & 78.5 & 64.3 & 68.6 & 77.8 & 63.2 & 69.8 \\
ResNeXt-101 & \textbf{64.8} & 76.5 & 48.6 & 57.3 & 79.3 & 66.6 & \textbf{70.2} & 78.5 & 65.4 & \textbf{71.3} \\
MnasNet & 53.8 & 61.8 & 42.9 & 49.9 & 73.0 & 59.4 & 63.0 & 72.1 & 58.1 & 64.3 \\
KFBNet & 58.8 & 68.8 & 45.2 & 53.3 & 77.9 & 64.2 & 68.1 & 77.3 & 63.0 & 69.4\\
FACNN & 56.5 & 60.3 & 48.7 & 52.6 & 73.1 & 65.3 & 66.8 & 71.6 & 64.1 & 67.7\\
SAFF & 61.8 & 72.5 & 48.1 & 56.7 & 79.4 & 63.9 & 68.6 & 78.7 & 62.8 & 69.9\\
LR-VGG-16 & 58.1 & 67.7 & 46.7 & 54.2 & 77.3 & 64.6 & 68.0 & 76.2 & 63.5 & 69.2 \\
LR-ResNet-50 & 63.1 & 68.1 & \textbf{53.1} & \textbf{59.0} & 76.7 & \textbf{67.6} & 69.7 & 75.3 & \textbf{66.5} & 70.6 \\
\Xhline{3\arrayrulewidth}
\end{tabular}
\end{table*}

\begin{sidewaystable*}[htbp!]
\centering
\renewcommand{\arraystretch}{0.9}
\caption{Comparisons of APs on the MultiScene-Clean dataset (\%). The best APs are shown in bold.}
\label{tab:msrlite_perclass_results}
\begin{tabular}{p{1.7cm}@{\hspace{0.05cm}}|*{36}{@{\hspace{0.06cm}}p{0.2cm}@{\hspace{0.37cm}}}}
\Xhline{3\arrayrulewidth}
{\footnotesize \textbf{Model}} & \rotatebox{50}{{\footnotesize apron}} & \rotatebox{50}{{\footnotesize baseball field}} & \rotatebox{50}{{\footnotesize basketball field}} & \rotatebox{50}{{\footnotesize beach}}& \rotatebox{50}{{\footnotesize bridge}} & \rotatebox{50}{{\footnotesize cemetery}} & \rotatebox{50}{{\footnotesize commercial}} & \rotatebox{50}{{\footnotesize farmland}}& \rotatebox{50}{{\footnotesize woodland}}& \rotatebox{50}{{\footnotesize golf course}}& \rotatebox{50}{{\footnotesize greenhouse}}& \rotatebox{50}{{\footnotesize helipad}}& \rotatebox{50}{{\footnotesize lake/pond}}& \rotatebox{50}{{\footnotesize oil field}}& \rotatebox{50}{{\footnotesize orchard}}& \rotatebox{50}{{\footnotesize parking lot}}& \rotatebox{50}{{\footnotesize park}}& \rotatebox{50}{{\footnotesize pier}}& \rotatebox{50}{{\footnotesize port}}& \rotatebox{50}{{\footnotesize quarry}}& \rotatebox{50}{{\footnotesize railway}}& \rotatebox{50}{{\footnotesize residential}}& \rotatebox{50}{{\footnotesize river}}& \rotatebox{50}{{\footnotesize roundabout}}& \rotatebox{50}{{\footnotesize runway}}& \rotatebox{50}{{\footnotesize soccer field}}& \rotatebox{50}{{\footnotesize solar farm}}& \rotatebox{50}{{\footnotesize sparse shrub}}& \rotatebox{50}{{\footnotesize stadium}}& \rotatebox{50}{{\footnotesize storage tanks}}& \rotatebox{50}{{\footnotesize tennis court}}& \rotatebox{50}{{\footnotesize train station}}& \rotatebox{50}{{\footnotesize wastewater plant}}& \rotatebox{50}{{\footnotesize wind turbine}}& \rotatebox{50}{{\footnotesize works}}& \rotatebox{50}{{\footnotesize sea}}\\
\hline

{\footnotesize SVM} & {\footnotesize 3.1} & {\footnotesize 6.3} & {\footnotesize 5.4} & {\footnotesize 4.0} & {\footnotesize 29.9} & {\footnotesize 4.5} & {\footnotesize 60.5} & {\footnotesize 44.1} & {\footnotesize 55.2} & {\footnotesize 3.1} & {\footnotesize 5.2} & {\footnotesize 3.0} & {\footnotesize 18.0} & {\footnotesize 0.1} & {\footnotesize 9.9} & {\footnotesize 65.5} & {\footnotesize 20.2} & {\footnotesize 5.5} & {\footnotesize 0.6} & {\footnotesize 3.0} & {\footnotesize 13.4} & {\footnotesize 68.3} & {\footnotesize 19.3} & {\footnotesize 8.7} & {\footnotesize 3.7} & {\footnotesize 9.5} & {\footnotesize 8.0} & {\footnotesize 25.7} & {\footnotesize 3.7} & {\footnotesize 5.2} & {\footnotesize 9.1} & {\footnotesize 2.7} & {\footnotesize 3.6} & {\footnotesize 0.8} & {\footnotesize 5.2} & {\footnotesize 3.4}\\

{\footnotesize RF} & {\footnotesize 3.1} & {\footnotesize 6.3} & {\footnotesize 5.4} & {\footnotesize 10.9} & {\footnotesize 28.1} & {\footnotesize 4.5} & {\footnotesize 62.7} & {\footnotesize 53.2} & {\footnotesize 54.6} & {\footnotesize 3.1} & {\footnotesize 5.4} & {\footnotesize 3.0} & {\footnotesize 17.4} & {\footnotesize 0.1} & {\footnotesize 10.1} & {\footnotesize 65.6} & {\footnotesize 20.2} & {\footnotesize 7.1} & {\footnotesize 0.6} & {\footnotesize 3.0} & {\footnotesize 13.6} & {\footnotesize 73.3} & {\footnotesize 19.5} & {\footnotesize 8.7} & {\footnotesize 3.7} & {\footnotesize 9.5} & {\footnotesize 10.6} & {\footnotesize 26.2} & {\footnotesize 3.7} & {\footnotesize 4.9} & {\footnotesize 9.1} & {\footnotesize 2.7} & {\footnotesize 3.6} & {\footnotesize 0.8} & {\footnotesize 5.2} & {\footnotesize 3.4}\\
         
{\footnotesize XGBOOST} & {\footnotesize 3.1} & {\footnotesize 9.4} & {\footnotesize 5.4} & {\footnotesize 15.8} & {\footnotesize 33.4} & {\footnotesize 4.5} & {\footnotesize 62.8} & {\footnotesize 56.2} & {\footnotesize 57.6} & {\footnotesize 3.1} & {\footnotesize 5.2} & {\footnotesize 3.0} & {\footnotesize 21.5} & {\footnotesize 0.1} & {\footnotesize 11.5} & {\footnotesize 68.8} & {\footnotesize 23.3} & {\footnotesize 9.3} & {\footnotesize 0.6} & {\footnotesize 3.0} & {\footnotesize 13.8} & {\footnotesize 74.7} & {\footnotesize 20.3} & {\footnotesize 8.7} & {\footnotesize 4.2} & {\footnotesize 9.8} & {\footnotesize 10.9} & {\footnotesize 32.9} & {\footnotesize 3.7} & {\footnotesize 5.2} & {\footnotesize 9.1} & {\footnotesize 2.7} & {\footnotesize 3.6} & {\footnotesize 0.8} & {\footnotesize 5.2} & {\footnotesize 4.3}\\

{\footnotesize VGG-16} & {\footnotesize 72.2} & {\footnotesize 81.7} & {\footnotesize 24.2} & {\footnotesize 70.0} & {\footnotesize 72.1} & {\footnotesize 28.9} & {\footnotesize 81.6} & {\footnotesize 87.8} & {\footnotesize 85.7} & {\footnotesize 65.1} & {\footnotesize 42.8} & {\footnotesize 34.9} & {\footnotesize 63.2} & {\footnotesize 1.9} & {\footnotesize 72.0} & {\footnotesize 86.2} & {\footnotesize 50.1} & {\footnotesize 72.7} & {\footnotesize 19.0} & {\footnotesize 50.9} & {\footnotesize 55.6} & {\footnotesize 93.7} & {\footnotesize 52.5} & {\footnotesize 65.8} & {\footnotesize 68.7} & {\footnotesize 61.6} & {\footnotesize 32.4} & {\footnotesize 59.7} & {\footnotesize 53.7} & {\footnotesize 49.4} & {\footnotesize 63.6} & {\footnotesize 35.5} & {\footnotesize 45.9} & {\footnotesize 51.6} & {\footnotesize 27.2} & {\footnotesize 54.9}\\

{\footnotesize VGG-19} & {\footnotesize 70.1} & {\footnotesize 80.7} & {\footnotesize 21.3} & {\footnotesize 67.1} & {\footnotesize 71.7} & {\footnotesize 28.0} & {\footnotesize 80.6} & {\footnotesize 87.4.} & {\footnotesize 85.5} & {\footnotesize 64.5} & {\footnotesize 44.4} & {\footnotesize 33.5} & {\footnotesize 64.1} & {\footnotesize 2.6} & {\footnotesize 73.4} & {\footnotesize 86.6} & {\footnotesize 50.7} & {\footnotesize 72.4} & {\footnotesize 17.2} & {\footnotesize 50.7} & {\footnotesize 56.4} & {\footnotesize 93.8} & {\footnotesize 52.3} & {\footnotesize 68.7} & {\footnotesize 68.6} & {\footnotesize 62.3} & {\footnotesize 32.3} & {\footnotesize 58.2} & {\footnotesize 53.0} & {\footnotesize 49.1} & {\footnotesize 66.8} & {\footnotesize 37.6} & {\footnotesize 47.0} & {\footnotesize 50.3} & {\footnotesize 28.3} & {\footnotesize 54.5}\\

{\footnotesize Inception-V3} & {\footnotesize 67.8} & {\footnotesize 83.2} & {\footnotesize 20.0} & {\footnotesize 68.8} & {\footnotesize 69.9} & {\footnotesize 19.8} & {\footnotesize 81.4} & {\footnotesize 85.0} & {\footnotesize 83.1} & {\footnotesize 51.4} & {\footnotesize 36.4} & {\footnotesize 33.0} & {\footnotesize 55.2} & {\footnotesize 0.5} & {\footnotesize 68.3} & {\footnotesize 86.0} & {\footnotesize 50.0} & {\footnotesize 70.5} & {\footnotesize 13.5} & {\footnotesize 49.0} & {\footnotesize 47.9} & {\footnotesize 92.3} & {\footnotesize 49.8} & {\footnotesize 63.9} & {\footnotesize 65.4} & {\footnotesize 59.7} & {\footnotesize 31.1} & {\footnotesize 56.6} & {\footnotesize 57.9} & {\footnotesize 44.6} & {\footnotesize 55.5} & {\footnotesize 34.5} & {\footnotesize 46.9} & {\footnotesize 44.5} & {\footnotesize 22.7} & {\footnotesize 59.6}\\

{\footnotesize ResNet-50} & {\footnotesize 77.3} & {\footnotesize 86.7} & {\footnotesize 26.4} & {\footnotesize 79.4} & {\footnotesize 74.6} & {\footnotesize 39.7} & {\footnotesize 83.3} & {\footnotesize 88.4} & {\footnotesize 86.7} & {\footnotesize 76.7} & {\footnotesize 49.3} & {\footnotesize 43.1} & {\footnotesize 66.0} & {\footnotesize 0.5} & {\footnotesize 76.8} & {\footnotesize 88.2} & {\footnotesize 55.3} & {\footnotesize 77.2} & {\footnotesize 24.7} & {\footnotesize 55.7} & {\footnotesize 62.4} & {\footnotesize 94.3} & {\footnotesize 59.6} & {\footnotesize 71.2} & {\footnotesize 74.8} & {\footnotesize 67.7} & {\footnotesize 40.3} & {\footnotesize 61.9} & {\footnotesize 63.0} & {\footnotesize 55.0} & {\footnotesize 68.8} & {\footnotesize 46.7} & {\footnotesize 54.5} & {\footnotesize 50.3} & {\footnotesize 36.0} & {\footnotesize 68.9}\\

{\footnotesize ResNet-101} & {\footnotesize 79.7} & {\footnotesize 88.0} & {\footnotesize 27.6} & {\footnotesize 80.2} & {\footnotesize 75.9} & {\footnotesize 44.5} & {\footnotesize 84.2} & {\footnotesize 88.4} & {\footnotesize 87.3} & {\footnotesize 75.6} & {\footnotesize 49.7} & {\footnotesize 45.3} & {\footnotesize 68.7} & {\footnotesize 0.9} & {\footnotesize 77.8} & {\footnotesize 88.6} & {\footnotesize 58.3} & {\footnotesize 77.6} & {\footnotesize 20.9} & {\footnotesize 61.2} & {\footnotesize 62.7} & {\footnotesize 94.5} & {\footnotesize 61.5} & {\footnotesize 73.3} & {\footnotesize 77.4} & {\footnotesize 70.0} & {\footnotesize 40.0} & {\footnotesize 62.1} & {\footnotesize 64.6} & {\footnotesize 54.2} & {\footnotesize 70.9} & {\footnotesize 46.6} & {\footnotesize 55.8} & {\footnotesize 47.2} & {\footnotesize \textbf{37.8}} & {\footnotesize 68.1}\\

{\footnotesize ResNet-152} & {\footnotesize 79.1} & {\footnotesize \textbf{88.6}} & {\footnotesize 27.4} & {\footnotesize \textbf{84.0}} & {\footnotesize 77.1} & {\footnotesize 42.4} & {\footnotesize 83.9} & {\footnotesize 88.7} & {\footnotesize 87.6} & {\footnotesize 77.4} & {\footnotesize 51.8} & {\footnotesize 46.9} & {\footnotesize 68.8} & {\footnotesize 0.4} & {\footnotesize 78.3} & {\footnotesize 89.2} & {\footnotesize \textbf{59.4}} & {\footnotesize \textbf{79.3}} & {\footnotesize 20.3} & {\footnotesize 59.4} & {\footnotesize 65.1} & {\footnotesize 94.5} & {\footnotesize 61.9} & {\footnotesize 74.4} & {\footnotesize 77.7} & {\footnotesize 70.7} & {\footnotesize 41.0} & {\footnotesize 62.8} & {\footnotesize 65.1} & {\footnotesize \textbf{57.2}} & {\footnotesize 72.5} & {\footnotesize 51.3} & {\footnotesize 57.3} & {\footnotesize 48.8} & {\footnotesize 35.5} & {\footnotesize \textbf{71.1}}\\

{\footnotesize SqueezeNet} & {\footnotesize 51.4} & {\footnotesize 73.8} & {\footnotesize 18.2} & {\footnotesize 57.6} & {\footnotesize 59.4} & {\footnotesize 15.6} & {\footnotesize 77.4} & {\footnotesize 83.8} & {\footnotesize 81.6} & {\footnotesize 50.1} & {\footnotesize 34.0} & {\footnotesize 12.3} & {\footnotesize 51.3} & {\footnotesize 0.7} & {\footnotesize 65.9} & {\footnotesize 83.7} & {\footnotesize 44.2} & {\footnotesize 56.8} & {\footnotesize 20.5} & {\footnotesize 34.7} & {\footnotesize 46.2} & {\footnotesize 93.2} & {\footnotesize 44.9} & {\footnotesize 57.7} & {\footnotesize 58.0} & {\footnotesize 45.7} & {\footnotesize 27.4} & {\footnotesize 53.2} & {\footnotesize 38.9} & {\footnotesize 35.9} & {\footnotesize 56.1} & {\footnotesize 22.8} & {\footnotesize 26.8} & {\footnotesize 18.0} & {\footnotesize 18.9} & {\footnotesize 49.1}\\

{\footnotesize MobileNet-V2} & {\footnotesize 74.3} & {\footnotesize 84.4} & {\footnotesize 24.8} & {\footnotesize 78.5} & {\footnotesize 73.4} & {\footnotesize 32.5} & {\footnotesize 81.9} & {\footnotesize 87.2} & {\footnotesize 85.9} & {\footnotesize 72.2} & {\footnotesize 46.3} & {\footnotesize 38.9} & {\footnotesize 64.3} & {\footnotesize 1.6} & {\footnotesize 73.8} & {\footnotesize 88.3} & {\footnotesize 53.7} & {\footnotesize 72.2} & {\footnotesize 18.1} & {\footnotesize 51.8} & {\footnotesize 60.0} & {\footnotesize 93.6} & {\footnotesize 55.8} & {\footnotesize 71.7} & {\footnotesize 67.8} & {\footnotesize 64.1} & {\footnotesize 34.3} & {\footnotesize 60.4} & {\footnotesize 58.7} & {\footnotesize 46.3} & {\footnotesize 69.8} & {\footnotesize 40.7} & {\footnotesize 47.1} & {\footnotesize 48.5} & {\footnotesize 31.5} & {\footnotesize 64.0} \\

{\footnotesize ShuffleNet-V2} & {\footnotesize 62.0} & {\footnotesize 78.4} & {\footnotesize 23.5} & {\footnotesize 67.5} & {\footnotesize 66.8} & {\footnotesize 14.4} & {\footnotesize 80.5} & {\footnotesize 84.1} & {\footnotesize 82.5} & {\footnotesize 61.0} & {\footnotesize 36.6} & {\footnotesize 18.0} & {\footnotesize 56.5} & {\footnotesize 0.4} & {\footnotesize 65.0} & {\footnotesize 85.5} & {\footnotesize 48.6} & {\footnotesize 61.9} & {\footnotesize 11.0} & {\footnotesize 40.4} & {\footnotesize 50.6} & {\footnotesize 92.5} & {\footnotesize 48.9} & {\footnotesize 57.0} & {\footnotesize 66.5} & {\footnotesize 59.4} & {\footnotesize 31.0} & {\footnotesize 57.3} & {\footnotesize 51.7} & {\footnotesize 37.5} & {\footnotesize 56.3} & {\footnotesize 30.2} & {\footnotesize 36.4} & {\footnotesize 20.4} & {\footnotesize 27.0} & {\footnotesize 58.8}\\

{\footnotesize DenseNet-121} & {\footnotesize 79.0} & {\footnotesize 87.5} & {\footnotesize 28.3} & {\footnotesize 80.9} & {\footnotesize 75.1} & {\footnotesize 37.9} & {\footnotesize 83.5} & {\footnotesize 87.7} & {\footnotesize 85.5} & {\footnotesize \textbf{78.0}} & {\footnotesize 48.0} & {\footnotesize 47.3} & {\footnotesize 66.3} & {\footnotesize 2.9} & {\footnotesize 75.7} & {\footnotesize 88.5} & {\footnotesize 58.3} & {\footnotesize 77.7} & {\footnotesize 25.3} & {\footnotesize 58.4} & {\footnotesize 62.3} & {\footnotesize 94.1} & {\footnotesize 58.5} & {\footnotesize 73.6} & {\footnotesize 73.9} & {\footnotesize 67.1} & {\footnotesize 39.1} & {\footnotesize 61.6} & {\footnotesize 61.4} & {\footnotesize 54.3} & {\footnotesize 69.9} & {\footnotesize 47.4} & {\footnotesize 56.4} & {\footnotesize 50.2} & {\footnotesize 31.0} & {\footnotesize 66.5}\\

{\footnotesize DenseNet-169} & {\footnotesize 81.8} & {\footnotesize 88.1} & {\footnotesize 26.3} & {\footnotesize 81.4} & {\footnotesize 76.9} & {\footnotesize 42.6} & {\footnotesize \textbf{84.5}} & {\footnotesize 88.1} & {\footnotesize 86.0} & {\footnotesize 77.2} & {\footnotesize 49.0} & {\footnotesize 44.1} & {\footnotesize 67.4} & {\footnotesize 4.3} & {\footnotesize 78.7} & {\footnotesize 88.8} & {\footnotesize 56.6} & {\footnotesize 78.1} & {\footnotesize 21.4} & {\footnotesize 58.2} & {\footnotesize 62.9} & {\footnotesize 94.4} & {\footnotesize 60.5} & {\footnotesize 72.4} & {\footnotesize 75.4} & {\footnotesize 69.4} & {\footnotesize 41.2} & {\footnotesize 61.2} & {\footnotesize \textbf{65.9}} & {\footnotesize 54.6} & {\footnotesize 72.4} & {\footnotesize 51.4} & {\footnotesize 58.6} & {\footnotesize 53.2} & {\footnotesize 35.1} & {\footnotesize 67.6}\\

{\footnotesize ResNeXt-50} & {\footnotesize 81.5} & {\footnotesize 87.1} & {\footnotesize 27.6} & {\footnotesize 81.6} & {\footnotesize 75.5} & {\footnotesize 41.3} & {\footnotesize 83.4} & {\footnotesize 88.2} & {\footnotesize 86.3} & {\footnotesize 76.1} & {\footnotesize 50.8} & {\footnotesize 50.3} & {\footnotesize 66.3} & {\footnotesize 9.6} & {\footnotesize 75.3} & {\footnotesize 89.0} & {\footnotesize 57.6} & {\footnotesize 77.7} & {\footnotesize 23.8} & {\footnotesize 58.6} & {\footnotesize 64.5} & {\footnotesize 94.1} & {\footnotesize 61.4} & {\footnotesize 73.3} & {\footnotesize 76.7} & {\footnotesize 68.7} & {\footnotesize 40.9} & {\footnotesize 62.2} & {\footnotesize 63.5} & {\footnotesize 53.8} & {\footnotesize 71.8} & {\footnotesize 50.2} & {\footnotesize 56.9} & {\footnotesize \textbf{54.7}} & {\footnotesize 34.9} & {\footnotesize 66.6}\\

{\footnotesize ResNeXt-101} & {\footnotesize \textbf{82.3}} & {\footnotesize 87.7} & {\footnotesize \textbf{30.2}} & {\footnotesize 82.9} & {\footnotesize \textbf{77.2}} & {\footnotesize \textbf{45.6}} & {\footnotesize 84.1} & {\footnotesize \textbf{88.8}} & {\footnotesize 87.3} & {\footnotesize 77.1} & {\footnotesize \textbf{52.6}} & {\footnotesize \textbf{54.8}} & {\footnotesize \textbf{71.0}} & {\footnotesize 1.3} & {\footnotesize \textbf{79.4}} & {\footnotesize \textbf{89.7}} & {\footnotesize 58.6} & {\footnotesize 76.1} & {\footnotesize 20.3} & {\footnotesize \textbf{61.8}} & {\footnotesize \textbf{66.5}} & {\footnotesize 94.5} & {\footnotesize \textbf{64.3}} & {\footnotesize 75.0} & {\footnotesize \textbf{79.2}} & {\footnotesize \textbf{72.0}} & {\footnotesize \textbf{42.3}} & {\footnotesize 63.3} & {\footnotesize 64.5} & {\footnotesize 55.4} & {\footnotesize 75.1} & {\footnotesize \textbf{54.4}} & {\footnotesize \textbf{58.7}} & {\footnotesize 52.5} & {\footnotesize 37.8} & {\footnotesize 67.2}\\

{\footnotesize MnasNet} & {\footnotesize 69.4} & {\footnotesize 84.0} & {\footnotesize 21.6} & {\footnotesize 70.9} & {\footnotesize 67.4} & {\footnotesize 20.7} & {\footnotesize 78.9} & {\footnotesize 84.7} & {\footnotesize 82.4} & {\footnotesize 63.0} & {\footnotesize 42.0} & {\footnotesize 27.5} & {\footnotesize 58.3} & {\footnotesize 0.7} & {\footnotesize 70.5} & {\footnotesize 85.7} & {\footnotesize 49.1} & {\footnotesize 69.5} & {\footnotesize 18.1} & {\footnotesize 45.2} & {\footnotesize 51.6} & {\footnotesize 91.4} & {\footnotesize 48.0} & {\footnotesize 66.1} & {\footnotesize 66.4} & {\footnotesize 59.0} & {\footnotesize 33.3} & {\footnotesize 55.9} & {\footnotesize 53.2} & {\footnotesize 42.4} & {\footnotesize 61.5} & {\footnotesize 32.8} & {\footnotesize 42.6} & {\footnotesize 33.4} & {\footnotesize 25.6} & {\footnotesize 62.6}\\

{\footnotesize KFBNet} & {\footnotesize 68.4} & {\footnotesize 83.0} & {\footnotesize 27.2} & {\footnotesize 75.1} & {\footnotesize 75.4} & {\footnotesize 37.6} & {\footnotesize 82.2} & {\footnotesize 88.7} & {\footnotesize 86.7} & {\footnotesize 68.4} & {\footnotesize 47.8} & {\footnotesize 47.0} & {\footnotesize 67.4} & {\footnotesize 6.3} & {\footnotesize 75.7} & {\footnotesize 89.1} & {\footnotesize 55.3} & {\footnotesize 73.1} & {\footnotesize 10.1} & {\footnotesize 57.2} & {\footnotesize 58.5} & {\footnotesize \textbf{94.6}} & {\footnotesize 55.1} & {\footnotesize 72.5} & {\footnotesize 68.7} & {\footnotesize 64.7} & {\footnotesize 35.4} & {\footnotesize 60.3} & {\footnotesize 46.5} & {\footnotesize 48.5} & {\footnotesize 74.8} & {\footnotesize 31.6} & {\footnotesize 47.3} & {\footnotesize 49.1} & {\footnotesize 26.0} & {\footnotesize 61.2}\\

{\footnotesize FACNN} & {\footnotesize 69.4} & {\footnotesize 83.7} & {\footnotesize 21.5} & {\footnotesize 70.5} & {\footnotesize 72.9} & {\footnotesize 31.7} & {\footnotesize 81.8} & {\footnotesize 88.4} & {\footnotesize 85.4} & {\footnotesize 66.4} & {\footnotesize 36.9} & {\footnotesize 36.2} & {\footnotesize 65.9} & {\footnotesize 4.8} & {\footnotesize 72.8} & {\footnotesize 87.5} & {\footnotesize 50.4} & {\footnotesize 71.1} & {\footnotesize 12.6} & {\footnotesize 57.8} & {\footnotesize 54.5} & {\footnotesize 93.6} & {\footnotesize 55.2} & {\footnotesize 70.1} & {\footnotesize 68.4} & {\footnotesize 64.5} & {\footnotesize 32.4} & {\footnotesize 56.4} & {\footnotesize 50.0} & {\footnotesize 50.8} & {\footnotesize 68.4} & {\footnotesize 32.3} & {\footnotesize 46.1} & {\footnotesize 46.4} & {\footnotesize 23.9} & {\footnotesize 54.2}\\

{\footnotesize SAFF} & {\footnotesize 74.5} & {\footnotesize 86.2} & {\footnotesize 29.8} & {\footnotesize 76.8} & {\footnotesize 75.1} & {\footnotesize 41.4} & {\footnotesize 83.0} & {\footnotesize 88.7} & {\footnotesize 86.6} & {\footnotesize 76.9} & {\footnotesize 50.0} & {\footnotesize 49.6} & {\footnotesize 67.7} & {\footnotesize 1.9} & {\footnotesize 76.8} & {\footnotesize 88.9} & {\footnotesize 58.7} & {\footnotesize 74.8} & {\footnotesize 17.1} & {\footnotesize 51.6} & {\footnotesize 62.1} & {\footnotesize 94.2} & {\footnotesize 58.8} & {\footnotesize \textbf{76.5}} & {\footnotesize 72.4} & {\footnotesize 68.3} & {\footnotesize 39.8} & {\footnotesize \textbf{63.3}} & {\footnotesize 57.6} & {\footnotesize 54.5} & {\footnotesize \textbf{77.8}} & {\footnotesize 41.9} & {\footnotesize 54.6} & {\footnotesize 46.6} & {\footnotesize 33.7} & {\footnotesize 65.9}\\

{\footnotesize LR-VGG-16} & {\footnotesize 76.3} & {\footnotesize 82.5} & {\footnotesize 19.9} & {\footnotesize 74.7} & {\footnotesize 71.0} & {\footnotesize 26.6} & {\footnotesize 82.5} & {\footnotesize 86.8} & {\footnotesize 86.4} & {\footnotesize 70.7} & {\footnotesize 41.4} & {\footnotesize 41.0} & {\footnotesize 65.1} & {\footnotesize \textbf{11.0}} & {\footnotesize 72.0} & {\footnotesize 87.5} & {\footnotesize 52.5} & {\footnotesize 73.1} & {\footnotesize 10.4} & {\footnotesize 57.4} & {\footnotesize 58.5} & {\footnotesize 93.5} & {\footnotesize 56.3} & {\footnotesize 70.7} & {\footnotesize 71.2} & {\footnotesize 62.7} & {\footnotesize 27.3} & {\footnotesize 58.1} & {\footnotesize 51.6} & {\footnotesize 50.5} & {\footnotesize 69.0} & {\footnotesize 43.9} & {\footnotesize 52.0} & {\footnotesize 45.9} & {\footnotesize 30.5} & {\footnotesize 62.5}\\

{\footnotesize LR-ResNet-50} & {\footnotesize 78.5} & {\footnotesize 88.2} & {\footnotesize 24.6} & {\footnotesize 80.8} & {\footnotesize 75.9} & {\footnotesize 44.2} & {\footnotesize 83.8} & {\footnotesize 88.7} & {\footnotesize \textbf{87.8}} & {\footnotesize 76.2} & {\footnotesize 50.9} & {\footnotesize 48.1} & {\footnotesize 67.4} & {\footnotesize 0.9} & {\footnotesize 77.1} & {\footnotesize 88.8} & {\footnotesize 57.3} & {\footnotesize 76.8} & {\footnotesize \textbf{29.3}} & {\footnotesize 57.5} & {\footnotesize 63.3} & {\footnotesize 94.2} & {\footnotesize 61.0} & {\footnotesize 72.4} & {\footnotesize 73.0} & {\footnotesize 70.5} & {\footnotesize 42.1} & {\footnotesize 62.3} & {\footnotesize 64.8} & {\footnotesize 55.1} & {\footnotesize 72.0} & {\footnotesize 48.7} & {\footnotesize 55.9} & {\footnotesize 48.9} & {\footnotesize 36.4} & {\footnotesize 67.3}\\

\Xhline{3\arrayrulewidth}
\end{tabular}
\end{sidewaystable*}

\begin{table*}
\caption{Example predictions of ResNext-101 on the MultiScene-Clean Dataset.}
\label{tab:example_predictions}
\centering
\renewcommand{\arraystretch}{1.3}
\begin{threeparttable}
\begin{tabular}{>{\centering\arraybackslash}m{2.5cm}>{\centering\arraybackslash}m{2.5cm}>{\centering\arraybackslash}m{2.5cm}>{\centering\arraybackslash}m{2.5cm}>{\centering\arraybackslash}m{2.5cm}>{\centering\arraybackslash}m{2.5cm}}
\hline
\Xhline{3\arrayrulewidth}
Multi-scene Aerial Images in the MultiScene-Clean dataset &
\vspace{0.5em}\includegraphics[width=0.14\textwidth]{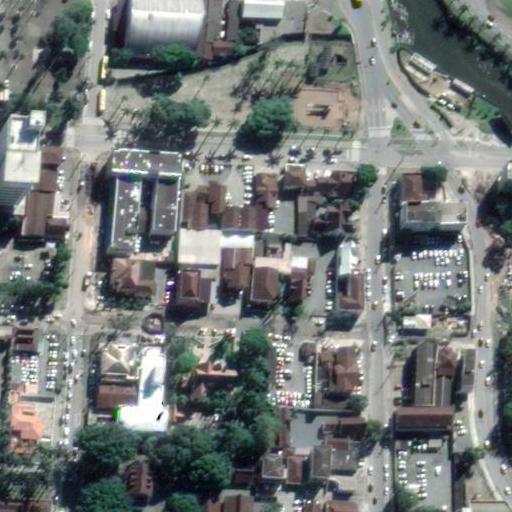} &
\vspace{0.5em}\includegraphics[width=0.14\textwidth]{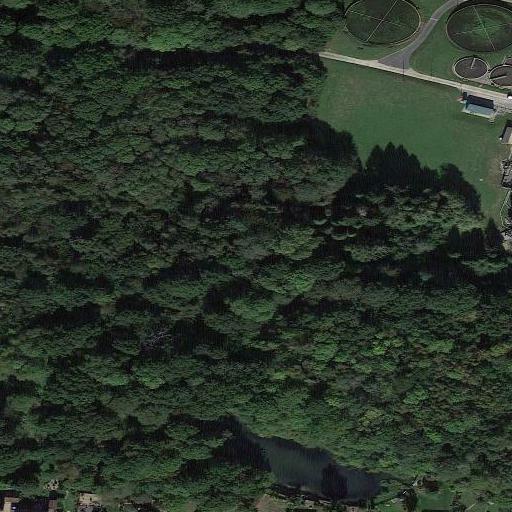} &
\vspace{0.5em}\includegraphics[width=0.14\textwidth]{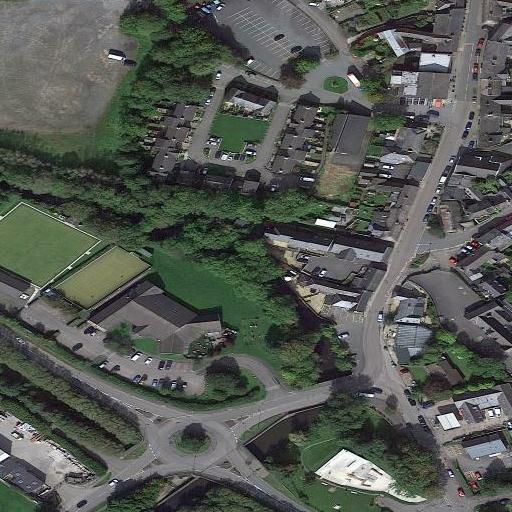}&
\vspace{0.5em}\includegraphics[width=0.14\textwidth]{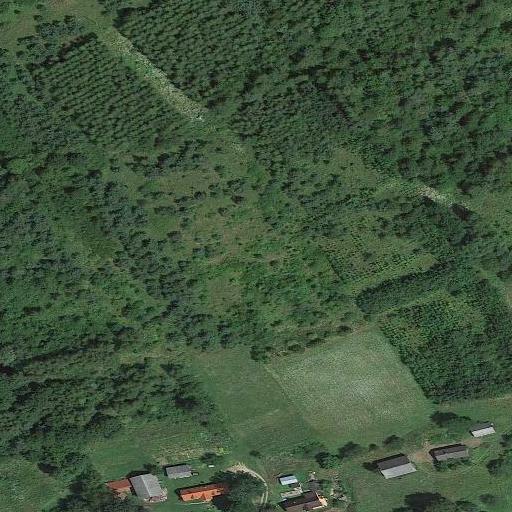}&
\vspace{0.5em}\includegraphics[width=0.14\textwidth]{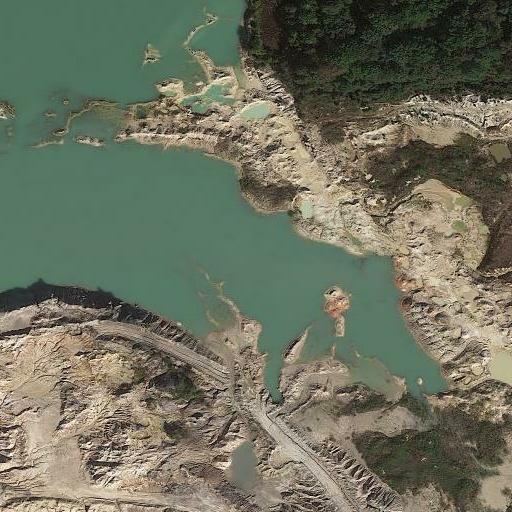} \\
\hline
Ground Truths & bridge, parking lot, river, roundabout, and residential &
woodland, lake/pond, and wastewater plant &
bridge, parking lot, river, roundabout, and residential &
farmland, woodland, orchard, residential, and sparse shrub &
lake/pond and quarry\\
\hline
Predictions & bridge, parking lot, river, roundabout, and residential &
woodland, lake/pond, and wastewater plant &
bridge, parking lot, river, roundabout, and residential &
farmland, woodland, orchard, residential, and sparse shrub &
lake/pond and quarry\\
\hline
Multi-scene Aerial Images in the MultiScene-Clean dataset &
\vspace{0.5em}\includegraphics[width=0.14\textwidth]{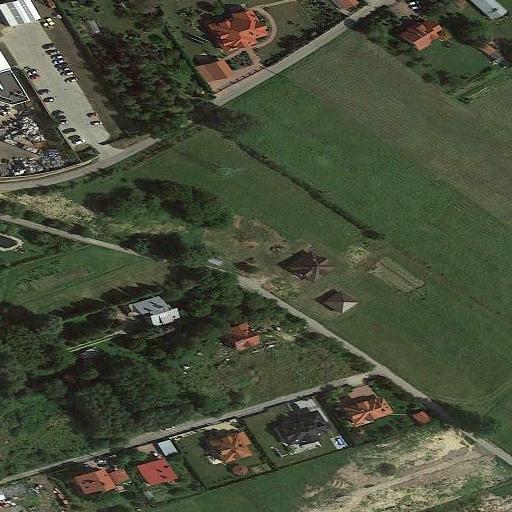} &
\vspace{0.5em}\includegraphics[width=0.14\textwidth]{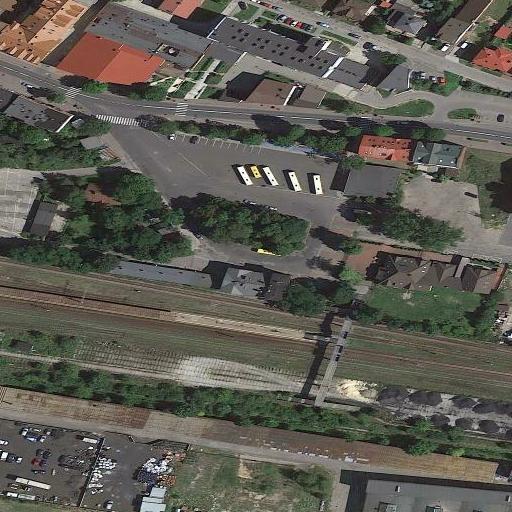} &
\vspace{0.5em}\includegraphics[width=0.14\textwidth]{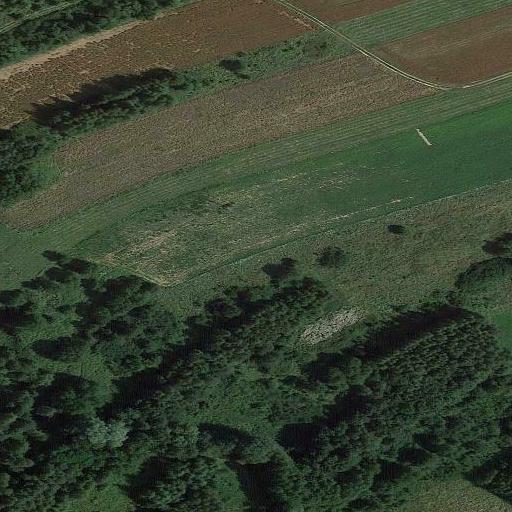} &
\vspace{0.5em}\includegraphics[width=0.14\textwidth]{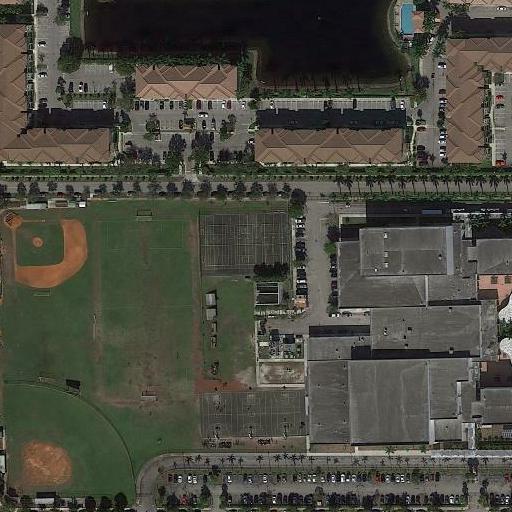} &
\vspace{0.5em}\includegraphics[width=0.14\textwidth]{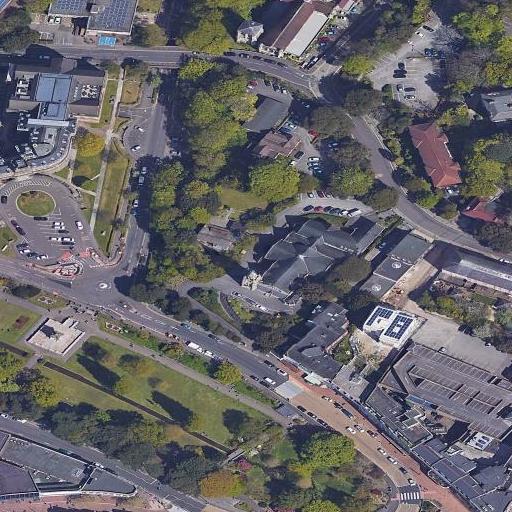} \\
\hline
Ground Truths & commercial, farmland, parking lot, and residential &
commercial, parking lot, park, railway, residential, train station, and works &
farmland, woodland, sparse shrub &
baseball field, basketball field, lake/pond, parking lot, residential, soccer field, and tennis court &
bridge, commercial, parking lot, park, residential, river, roundabout, and solar farm\\
\hline
Predictions & \textcolor{purple}{commercial}, farmland, \textcolor{blue}{woodland}, \textcolor{purple}{parking lot}, and residential &
\textcolor{purple}{commercial}, parking lot, \textcolor{purple}{park}, railway, residential, train station, and \textcolor{purple}{works}&
farmland, woodland, \textcolor{blue}{lake/pond}, and sparse shrub &
\textcolor{purple}{baseball field}, \textcolor{purple}{baseketball field}, lake/pond, \textcolor{purple}{parking lot}, residential, \textcolor{purple}{soccer field}, and tennis court &
bridge, commercial, \textcolor{blue}{lake/pond}, parking lot, \textcolor{purple}{park}, \textcolor{purple}{river}, \textcolor{purple}{solar farm}, and residential\\
\hline
\Xhline{3\arrayrulewidth}
\end{tabular}
\begin{tablenotes}
\item[] \textcolor{purple}{Purple} predictions indicate false negatives, while \textcolor{blue}{blue} predictions are false positives.
\end{tablenotes}
\end{threeparttable}
\end{table*}

\subsection{Baselines}

To provide comprehensive benchmarks, we evaluate the performance of extensive popular deep neural networks. Since they were originally designed for single-label classification, we substitute sigmoid functions for their softmax activations to predict multiple scene labels that are encoded into multi-hot binary sequences. Besides, several classical machine learning algorithms are also evaluated. In total, 22 models are tested on both MultiScene-Clean and MultiScene datasets, and a brief review is as follows.

\begin{itemize}

\item \textit{SVM~\cite{cortes1995svm}:} Support vector machine (SVM) aims to learn one or several hyperplanes for separating samples of different classes with the largest margin. Usually, the hyperplanes are constructed in a high dimensional space, and can be learned directly (Linear SVM) or through kernel functions (Nonlinear SVM). In our experiments, we select the latter and use a radial basis function (RBF) kernel~\cite{vert2004rbf} to learn SVM.

\item \textit{RF~\cite{ho1995rf}:} Random forest (RF) is an ensemble of decision trees, which are trained with random subspaces of image features and make final predictions through the majority voting. The number of decision trees is set to 200 in our experiments.

\item \textit{XGBOOST:} XGBOOST\footnote{https://xgboost.readthedocs.io/en/latest/tutorials/model.html} is a computationally efficient implementation of gradient-boosted trees~\cite{hastie2009gbtree} that optimizes tree ensembles (e.g., an ensemble of decision trees) through successive learning steps~\cite{friedman2001gb}. In each step, the existing trees are fixed, and a new tree is added and optimized with objective functions. Considering the difficulty of our task, we set the number of trees to 200 for training XGBOOST on both datasets.

\item \textit{VGGNet~\cite{vgg}:} VGGNet utilizes five convolutional blocks and three fully-connected layers to extract high-level features for image classification. Each block has multiple stacked convolutional layers and ends with one max-pooling layer. The size of convolutional filters is $3 \times 3$, and the stride of max-pooling layers is 2. In our experiments, a 16-layer VGGNet (VGG-16) and a 19-layer VGGNet (VGG-19) are trained on our dataset.

\item \textit{Inception networks~\cite{inceptionv1,inceptionv2,inceptionv3,inceptionv4}:} Inception networks are characterized by their wide modules, where convolutional filters of variant sizes and max-pooling operators are jointly employed to learn diverse features. Besides, a bottleneck architecture made of $1 \times 1$ convolutions is introduced to mitigate the boosted computational cost resulting from heavy inception modules. In Table~\ref{tab:msrlite_overall_results} and \ref{tab:msrlite_perclass_results}, we report the performance of Inception-v3~\cite{inceptionv3} in multi-scene recognition.

\item \textit{ResNet~\cite{resnet}:} ResNet aims to address the degradation problem by learning residual mappings with shortcut connections. By doing so, ResNet can go much deeper than plain CNNs and achieve outstanding performance in not only image classification but also semantic segmentation and object detection tasks. In our experiments, we evaluate a 50-layer ResNet (ResNet-50), a 101-layer ResNet (ResNet-101), and a 152-layer ResNet (ResNet-152) on the proposed dataset. Notably, residual blocks in these deep ResNets are modified into bottleneck architectures for reducing the computational burden.

\item \textit{SqueezeNet~\cite{squeezenet}:} SqueezeNet focuses on preserving network performance with fewer parameters. To achieve this, most of $3 \times 3$ convolutional filters are replaced with $1 \times 1$ filters, and features are squeezed in the channel dimension before fed into the remaining $3 \times 3$ filters. In addition, bypass connections are introduced to features of the same size for improving the classification performance. Experimental results of SqueezeNet on our dataset are reported in Section~\ref{sec:msrlite_results} and \ref{sec:msr_results}.

\item \textit{MobileNet~\cite{mobilenet}:} MobileNet is a light-weight deep neural network, which is applicable on mobile devices with restricted computational sources. The network is designed in a streamlined architecture, and depthwise separable convolutions play a significant role in increasing computational efficiency. Specifically, such convolutions are implemented by factorizing standard convolutions into depthwise and pointwise convolutions. The former is conducted on each channel, and the latter aggregates channel-wise outputs via $1 \times 1$ convolutions. To further reduce the computational cost, two hyperparameters, width multiplier $\alpha$ and resolution multiplier $\beta$, are designed to shrink feature channels and input resolutions, respectively. In the advanced variation of MobileNet, i.e., MobileNet-V2~\cite{mobilenetv2}, inverted residual connections and linear bottlenecks are developed to improve the network performance. In our experiments, we train MobileNet-V2 and set both $\alpha$ and $\beta$ as the default value, 1.

\item \textit{ShuffleNet~\cite{shufflenet}:} ShuffleNet improves computational efficiency by utilizing pointwise group convolutions and channel shuffle. Specifically, the former divides feature maps into several groups and conducts $1 \times 1$ convolutions on each group independently. The latter rearranges feature channels for enabling information to flow across channels belonging to different groups. Besides, element-wise addition, which is often used in a residual block, is replaced with concatenation for enlarging channel dimension at a low computational cost. In ShuffleNet-V2~\cite{shufflenetv2}, features are grouped by channel split, and pointwise group convolutions are discarded. As a consequence, two feature groups are yielded and fed into two branches, of which one is an identity mapping and the other is a set of convolutions. Afterwards, outputs are concatenated and shuffled along the channel dimension. In our experiments, we evaluate the performance of ShuffleNet-V2 on our dataset.

\item \textit{DenseNet~\cite{densenet}:} DenseNet proposes to enhance information flow by directly connecting each layer to all subsequent layers with equivalent feature-map sizes. To preserve information learned by proceeding layers, concatenation is employed to combine features from various layers. By reusing feature maps throughout entire networks, DenseNet can learn compact internal representations for visual recognition tasks. Two variations, a 121-layer DenseNet (DenseNet-121) and a 169-layer DenseNet (DenseNet-169), are tested.

\begin{table*}[t!]
\footnotesize
\centering
\renewcommand{\arraystretch}{1}
\caption{Numerical results of baseline models on the MultiScene dataset (\%). Models are trained on images with noisy crowdsourced annotations and tested on cleanly-labeled images. The best scores are shown in bold.}
\label{tab:msr_overall_results}
\begin{tabular}{p{2cm}|*{4}{p{0.6cm}}|*{3}{p{0.6cm}}|*{3}{p{0.6cm}}}
\Xhline{3\arrayrulewidth}
\textbf{Model} & \textbf{mAP} & \textbf{mCP} & \textbf{mCR} & \textbf{mCF$_1$} & \textbf{mEP} & \textbf{mER} & \textbf{mEF$_1$} & \textbf{OP} & \textbf{OR} & \textbf{OF$_1$} \\
\hline
SVM & 14.7 & 24.7 & 4.1 & 5.4 & 51.4 & 15.7 & 23.1 & 77.7 & 15.5 & 25.8 \\
RF & 15.1 & 49.7 & 4.4 & 6.1 & 55.4 & 16.4 & 34.3 & 78.7 & 15.8 & 26.3 \\
XGBOOST & 18.4 & 54.6 & 10.6 & 14.9 & 62.0 & 26.7 & 35.1 & 70.4 & 25.5 & 37.4 \\
VGG-16 & 63.4 & 71.0 & 46.9 & 54.1 & 78.4 & 51.6 & 59.3 & 79.3 & 49.6 & 61.0\\
VGG-19 & 59.8 & 68.9 & 47.2 & 54.1 & 75.5 & 52.2 & 58.9 & 75.1 & 50.2 & 60.2\\
Inception-V3 & 65.8 & 74.1 & 50.8 & 58.5 & 79.1 & 53.8 & 61.2 & 79.5 & 51.9 & 62.8\\
ResNet-50 & 63.9 & 73.7 & 47.7 & 55.9 & 78.3 & 52.5 & 60.0 & 78.5 & 50.7 & 61.6 \\
ResNet-101 & 63.4 & 73.0 & 47.5 & 55.5 & 77.1 & 52.5 & 59.7 & 77.2 & 50.6 & 61.2\\
ResNet-152 & 62.8 & 73.2 & 47.6 & 55.7 & 76.2 & 53.1 & 59.9 & 76.5 & 51.3 & 61.4\\
SqueezeNet & 61.4 & 74.4 & 41.1 & 50.5 & 78.9 & 47.7 & 56.4 & 80.7 & 45.9 & 58.5\\
MobileNet-V2 & 65.5 & 72.3 & 48.4 & 56.0 & 79.6 & 54.6 & 62.0 & 80.1 & 52.8 & 63.6 \\
ShuffleNet-V2 & 65.1 & 74.6 & 46.7 & 55.1 & 81.7 & 51.0 & 59.9 & \textbf{82.9} & 49.0 & 61.6\\
DenseNet-121 & 67.5 & 77.0 & 49.4 & 58.2 & \textbf{82.2} & 54.4 & \textbf{62.6} & 82.8 & 52.3 & \textbf{64.1}\\
DenseNet-169 & 64.2 & 71.3 & \textbf{53.3} & \textbf{59.3} & 77.2 & \textbf{55.7} & 62.0 & 77.1 & \textbf{53.9} & 63.4\\
ResNeXt-50 & 63.9 & 73.6 & 49.0 & 56.9 & 77.5 & 52.6 & 59.8 & 77.6 & 50.7 & 61.3\\
ResNeXt-101 & 60.8 & 68.5 & 47.4 & 53.7 & 73.8 & 51.2 & 57.7 & 74.0 & 49.5 & 59.3\\
MnasNet & 58.1 & 74.1 & 31.0 & 40.4 & 75.0 & 38.0 & 47.6 & 80.4 & 36.0 & 49.7\\
KFBNet & 67.1 & \textbf{77.7} & 46.2 & 54.3 & 80.2 & 54.0 & 61.7 & 81.1 & 52.3 & 63.6\\
FACNN & 65.2 & 73.9 & 47.6 & 55.6 & 78.9 & 53.9 & 61.1 & 80.0 & 52.1 & 63.1\\
SAFF & 64.8 & 74.0 & 47.4 & 55.3 & 80.9 & 51.2 & 59.8 & 81.8 & 49.4 & 61.6\\
LR-VGG-16 & \textbf{67.8} & 76.0 & 48.4 & 56.1 & 80.5 & 52.1 & 60.5 & 81.3 & 50.3 & 62.2\\
LR-ResNet-50 & 65.5 & 71.2 & 51.6 & 57.9 & 79.2 & 53.1 & 60.7 & 79.4 & 51.2 & 62.3\\
\Xhline{3\arrayrulewidth}
\end{tabular}
\end{table*}

\item \textit{ResNeXt~\cite{resnext}:} ResNeXt learns residuals with aggregated residual transformations but not a stack of convolutional layers (e.g., ResNet). The aggregated residual transformation is implemented by first slicing features into low-dimensional embeddings and then conducting convolutions on them. Afterwards, outputs are aggregated with element-wise addition. With this design, ResNeXt outperforms its ResNet counterpart on ImageNet-5K~\cite{resnext} and COCO~\cite{coco} datasets. We test a 50-layer ResNext (ResNeXt-50) and a 101-layer (ResNeXt-101) in our experiments.

\item \textit{MnasNet~\cite{mnasnet}:} MnasNet architectures are automatically learned on target datasets through a mobile neural architecture search (MNAS) algorithm~\cite{mnasnet}. Compared to conventional NAS algorithms~\cite{nas}, MNAS takes not only classification accuracy but also model latency into consideration and is executed on mobile phones for measuring real-world inference latency. As a consequence, MnasNet searched on target datasets is expected to achieve a good trade-off between accuracy and latency. To control the model size, a depth multiplier is designed for scaling the number of channels in each layer. In our experiments, the depth multiplier is set to 1, and the best-performing MnasNet searched on the ImageNet dataset~\cite{imagenet} is chosen to perform multi-scene recognition in the wild.

\item \textit{KFBNet~\cite{li2020kfb}:} KFBNet exploits a key region capturing method
namely key filter bank (KFB) for aerial image scene classification. The proposed KFB is composed of two streams: a global stream (G-Stream) and a key stream (K-Stream). The former predicts labels using features learned by the last block of a CNN, while the latter highlights key features in both spatial and channel dimensions for inferring scene categories. Finally, predictions made by the two streams are merged via an element-wise addition as the final decision. We take VGG-16 as the backbone and report numerical results in Table~\ref{tab:msrlite_overall_results}, \ref{tab:msrlite_perclass_results}, and \ref{tab:msr_overall_results}.

\item \textit{FACNN~\cite{lu2019faccnn}:} FACNN is a scene classification network composed of a CNN backbone and a Feature Aggregation module. In the latter, features extracted by the last three blocks of VGG-16 are aggregated through pooling operations and $1 \times 1$ convolutions. Afterwards, they are concatenated with outputs of the second fully-connected layer of VGG-16 to form discriminative scene representations for the final prediction.

\item \textit{SAFF~\cite{cao2020saff}:} SAFF proposes a non-parametric self-attention layer for enhancing spatial and channel responses of feature maps. Specifically, features extracted by the last three blocks of a pre-trained CNN (e.g., VGG-16) are fused and fed into the proposed self-attention layer. In this layer, spatial- and channel-wise weightings are conducted to emphasize the importance of locations of salient objects and channels with infrequently occurring features, respectively. Principal component analysis (PCA) whitening is also introduced to reduce the information redundancy and squash channels. However, since this operation frequently fails the network training, we replace it with a learnable fully-connected layer. Besides, VGG-16 is selected as the backbone in our experiments.

\item \textit{LR-CNN~\cite{hua2020relation}:} LR-CNN is a multi-label classification network, which consists of three elements: a class-wise feature extraction module, an attentional region extraction module, and a relational reasoning module. Specifically, the first module learns deep features with respect to each category from the input images. Afterwards, the second module extracts attentional regions of class-wise features, which are eventually leveraged to reason about relations between different objects for inferring their existences through the third module. In our experiments, we validate LR-VGG-16 and LR-ResNet-50, where VGG-16 and ResNet-50 are taken as backbones, respectively.

\end{itemize}

\begin{figure*}
\centering
\includegraphics[width=.95\textwidth]{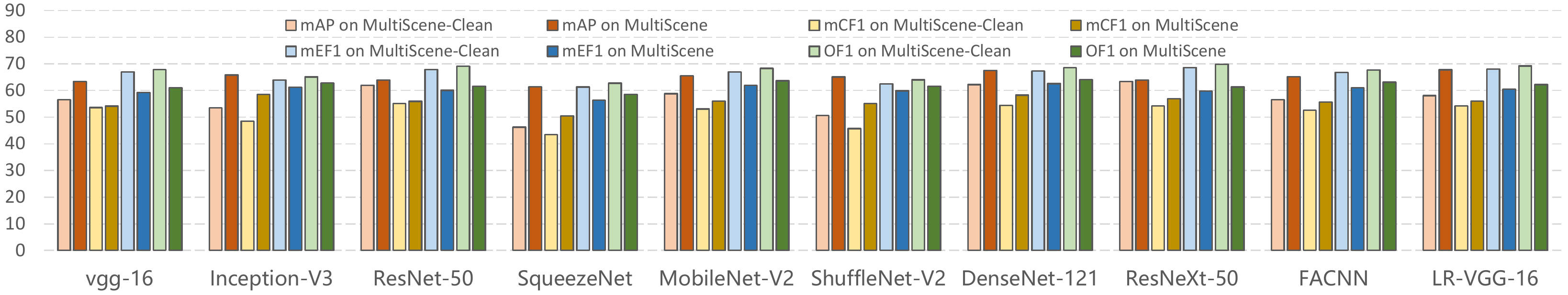}
\caption{Comparisons of the performance of networks trained on images with clean (light-color bars) and crowdsourced (dark-color bars) annotations, respectively. For each network, the left four bars represent class-based scores, mAPs and CF$_1$, while the right four bars indicate EF$_1$ and OF$_1$ scores.}
\label{fig:osm_gt}
\end{figure*}

\subsection{Training Details}
Before training SVM, RF, and XGBOOST, we use histogram of oriented gradient (HOG)~\cite{dalal2005hog} and local binary pattern (LBP)~\cite{ojala2002lbp} as visual features as recommended in \cite{lin2011large}. The size of each cell is set to $32 \times 32$ pixels for HOG, and the radius is defined as $16$ pixels for LBP. We use Scipy to implement these machine learning classifiers and apply them to multi-scene recognition using the function \textit{MultiOutputClassifier\footnote{https://scikit-learn.org/stable/modules/generated/sklearn.multioutput.Multi-OutputClassifier.html}}. As to baseline classification neural networks, we initialize them with weights pre-trained on the ImageNet dataset and fine-tune them on the proposed multi-scene image dataset. The loss is defined as binary cross-entropy, and stochastic gradient descent (SGD) with momentum~\cite{sgdm} is selected as the optimizer. To accelerate the network convergence, the momentum is set to a large value, 0.9. Besides, the initial learning rate and weight decay are set to 0.02 and $1e-4$, respectively. All deep networks are implemented on Pytorch and validated on one NVIDIA Tesla V100-SXM2 32GB GPU. For experiments on both MultiScene-Clean and MultiScene, we train networks for 87k and 581k iterations, respectively, and the size of each training batch is set to 16 for both versions.

\subsection{Experimental Results across Different Tasks}
\label{sec:msrlite_results}
\subsubsection{Multi-scene Recognition with Cleanly-labeled Data}
To evaluate baselines for our task, we conduct experiments on the MultiScene-Clean dataset and report quantitative results in Table~\ref{tab:msrlite_overall_results}. It can be seen that ResNeXt-101 achieves the best mAP (64.8\%), mEF$_1$ (70.2\%), and OF$_1$ score (71.3\%), which demonstrate its high performance and robustness in this task from almost all perspectives. LR-ResNet-50 gains the highest value in mCF$_1$ (59.0\%) owing to its capability of reasoning about relations among various scenes. Moreover, such a reasoning capability also enables LR-ResNet-50 to surpass the other baselines in all recall metrics, as scenes tend to be predicted as positive once its related scenes are recognized. Another observation is that MnasNet, SqueezeNet, and ShuffleNet-V2 show relatively poor performance due to their light-weight designs. Compared to deep neural networks, traditional machine learning algorithms achieve lower scores in all metrics.

For an insight into the performance of networks in identifying different scenes, we also report per-class APs in Table~\ref{tab:msrlite_perclass_results}. As we can see, ResNeXt-101 achieves the highest APs in most scenes, which is in line with the previous observations. Furthermore, we note that most networks fail to accurately recognize senes having scarce training samples, e.g., oil field and port. This suggests that learning unbiased models on an imbalanced dataset is a big challenge. Besides numerical results, we exhibit several predictions in Table~\ref{tab:example_predictions}.

\subsubsection{Learning from Noisy Crowdsourced Labels}
\label{sec:msr_results}

We investigate networks learned from noisy crowdsourced labels for our task on the MultiScene dataset. To ensure a fair comparison, we utilize the same test set as in Section~\ref{sec:msrlite_results} and report numerical results in Table~\ref{tab:msr_overall_results}. It can be observed that OF$_1$ scores of all models are decreased by an average of 8.2\% compared to the values in Table~\ref{tab:msrlite_overall_results}, which demonstrates that noise in crowdsourced annotations significantly affects the learning of deep neural networks. Moreover, it is interesting to note that the values of class-based metrics, mAP and mCF$_1$ score, are increased by 4.6\% and 1.2\%, respectively, in comparison with those in Table~\ref{tab:msrlite_overall_results}. This can be attributed to the fact that numbers of training samples, especially for scenes seldomly appearing, are effortlessly increased by crawling OSM data with keyword searching. Compared to models showing high performance on the MultiScene-Clean dataset, we find that DenseNet gains the highest scores in mCF$_1$ (59.3\%), mEF$_1$ (62.6\%), and mOF$_1$ (64.1\%), as it can sufficiently reuse features and has relatively few parameters. Besides, LR-VGG-16 achieves the highest mAP (67.8\%), which demonstrates that taking advantage of underlying relations among various scenes can suppress the influence of noise introduced by OSM data. Furthermore, we compare the performance of several networks trained on MultiScene-Clean and MultiScene datasets in Fig.~\ref{fig:osm_gt}, and it can be again observed that higher class-based scores (see \textcolor[RGB]{197, 90, 17}{orange} and \textcolor[RGB]{191, 144, 0}{brown} bars in Fig.~\ref{fig:osm_gt}) are obtained when using massive crowdsourced labels. All in all, although crowdsourced labels influence the overall performance of networks, comparisons in class-based scores also suggest their great potential.

\section{Conclusion}
\label{sec:conclusion}

In this paper, we propose a large-scale dataset, MultiScene, for multi-scene recognition in single images, which is featured by unconstrained multi-scene aerial images and the available both crowdsourced and clean labels. The proposed dataset allows researches in not only recognizing aerial scenes in the wild but also learning from noisy crowdsourced labels. We comprehensively evaluate popular baseline models on both MultiScene-Clean (a subset consisting of only cleanly-labeled images) and MultiScene datasets. Experimental results on the former demonstrate that unconstrained multi-scene recognition is still a challenging task, and those on the latter showcase the great potential of exploiting a large number of  crowdsourced annotations. Looking into the future, the dataset can be applied to develop more efficient networks and learning strategies for exploiting noisy labels for aerial scene understanding in the wild.

\ifCLASSOPTIONcaptionsoff
\newpage
\fi

\bibliographystyle{IEEEtran}
\bibliography{reference}
\begin{IEEEbiography}[{\includegraphics[width=1in,height=1.25in,clip,keepaspectratio]{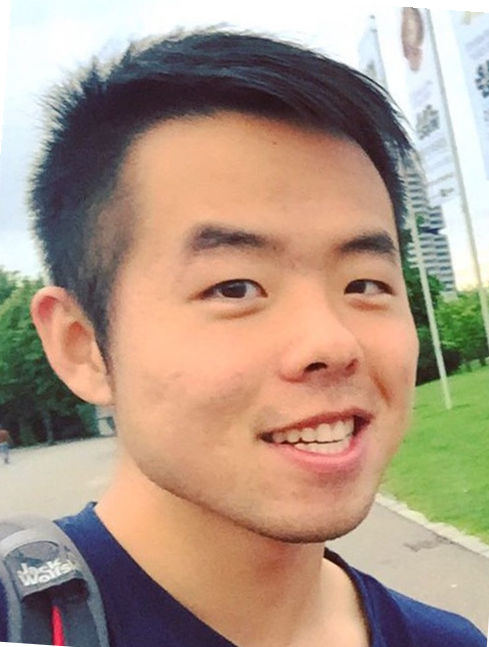}}]{Yuansheng Hua}
(S'18) received the bachelor's degree in remote sensing science and technology from the Wuhan University, Wuhan, China, in 2014, and double master's degrees in Earth Oriented Space Science and Technology (ESPACE) and Photogrammetry and remote sensing from the Technical University of Munich (TUM), Munich, Germany, and Wuhan University, Wuhan, China, in 2018 and 2019, respectively. He is currently pursuing the Ph.D. degree with the German Aerospace Center (DLR), Wessling, Germany and the Technical University of Munich (TUM), Munich, Germany. 

In 2019, he was a visiting researcher with the Wageningen University \& Research, Wageningen, Netherlands. His research interests include remote sensing, computer vision, and deep learning, especially their applications in remote sensing.

\end{IEEEbiography}

\begin{IEEEbiography}[{\includegraphics[width=1in,height=1.25in,clip,keepaspectratio]{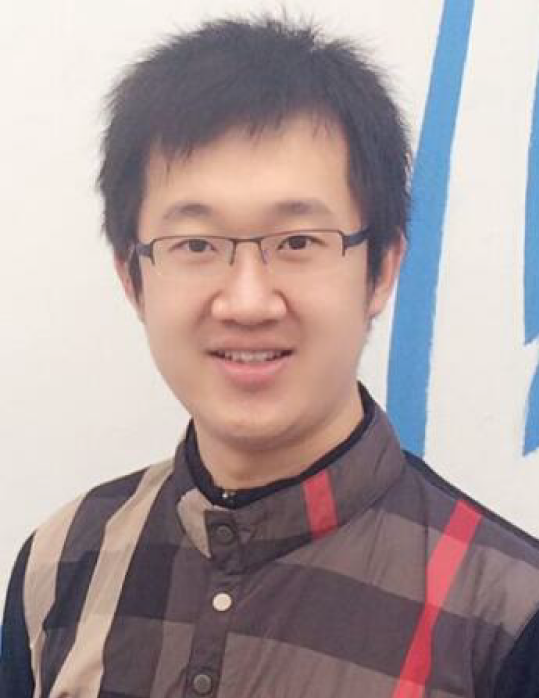}}]{Lichao Mou}
received the Bachelor's degree in automation from the Xi'an University of Posts and Telecommunications, Xi'an, China, in 2012, the Master's degree in signal and information processing from the University of Chinese Academy of Sciences (UCAS), China, in 2015, and the Dr.-Ing. degree from the Technical University of Munich (TUM), Munich, Germany, in 2020.
\par
He is currently a Guest Professor at the Munich AI Future Lab AI4EO, TUM and the Head of Visual Learning and Reasoning team at the Department ``EO Data Science'', Remote Sensing Technology Institute (IMF), German Aerospace Center (DLR), Wessling, Germany. Since 2019, he is a Research Scientist at DLR-IMF and an AI Consultant for the Helmholtz Artificial Intelligence Cooperation Unit (HAICU). In 2015 he spent six months at the Computer Vision Group at the University of Freiburg in Germany. In 2019 he was a Visiting Researcher with the Cambridge Image Analysis Group (CIA), University of Cambridge, UK.
\par
He was the recipient of the first place in the 2016 IEEE GRSS Data Fusion Contest and finalists for the Best Student Paper Award at the 2017 Joint Urban Remote Sensing Event and 2019 Joint Urban Remote Sensing Event.
\end{IEEEbiography}


\begin{IEEEbiography}[{\includegraphics[width=1in,height=1.25in,clip,keepaspectratio]{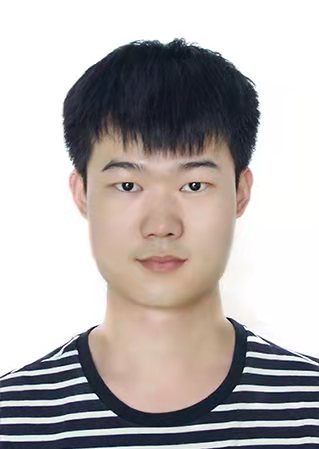}}]{Pu Jin}
(S'21) received the bachelor's degree in electronic information science and technology from the Wuhan University, Wuhan, China, in 2017, and double master's degrees in Earth Oriented Space Science and Technology (ESPACE) and Photogrammetry and remote sensing from the Technical University of Munich (TUM), Munich, Germany, and Wuhan University, Wuhan, China, in 2020 and 2021, respectively. He is currently pursuing the Ph.D. degree with the German Aerospace Center (DLR), Wessling, Germany and the Technical University of Munich (TUM), Munich, Germany.
His research interests include remote sensing, computer vision, and deep learning, especially their applications in remote sensing.
\end{IEEEbiography}

\begin{IEEEbiography}[{\includegraphics[width=1in,height=1.25in,clip,keepaspectratio]{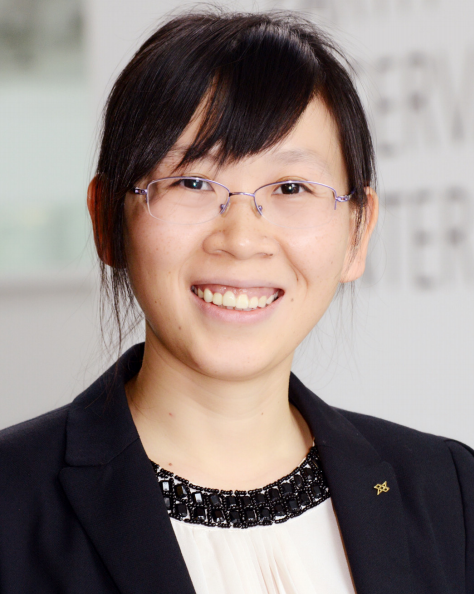}}]{Xiao Xiang Zhu}(S'10--M'12--SM'14--F'21) received the Master (M.Sc.) degree, her doctor of engineering (Dr.-Ing.) degree and her “Habilitation” in the field of signal processing from Technical University of Munich (TUM), Munich, Germany, in 2008, 2011 and 2013, respectively.
\par
She is currently the Professor for Data Science in Earth Observation (former: Signal Processing in Earth Observation) at Technical University of Munich (TUM) and the Head of the Department ``EO Data Science'' at the Remote Sensing Technology Institute, German Aerospace Center (DLR). Since 2019, Zhu is a co-coordinator of the Munich Data Science Research School (www.mu-ds.de). Since 2019 She also heads the Helmholtz Artificial Intelligence -- Research Field ``Aeronautics, Space and Transport". Since May 2020, she is the director of the international future AI lab "AI4EO -- Artificial Intelligence for Earth Observation: Reasoning, Uncertainties, Ethics and Beyond", Munich, Germany. Since October 2020, she also serves as a co-director of the Munich Data Science Institute (MDSI), TUM. Prof. Zhu was a guest scientist or visiting professor at the Italian National Research Council (CNR-IREA), Naples, Italy, Fudan University, Shanghai, China, the University  of Tokyo, Tokyo, Japan and University of California, Los Angeles, United States in 2009, 2014, 2015 and 2016, respectively. She is currently a visiting AI professor at ESA's Phi-lab. Her main research interests are remote sensing and Earth observation, signal processing, machine learning and data science, with a special application focus on global urban mapping.

Dr. Zhu is a member of young academy (Junge Akademie/Junges Kolleg) at the Berlin-Brandenburg Academy of Sciences and Humanities and the German National  Academy of Sciences Leopoldina and the Bavarian Academy of Sciences and Humanities. She serves in the scientific advisory board in several research organizations, among others the German Research Center for Geosciences (GFZ) and Potsdam Institute for Climate Impact Research (PIK). She is an associate Editor of IEEE Transactions on Geoscience and Remote Sensing and serves as the area editor responsible for special issues of IEEE Signal Processing Magazine. She is a Fellow of IEEE.
\end{IEEEbiography}
\end{document}